%% file: neurips_2024.tex
\documentclass{article}

\usepackage[preprint,nonatbib]{neurips_2024}

\usepackage[utf8]{inputenc} 
\usepackage[T1]{fontenc}    
\usepackage{hyperref}       
\usepackage{url}            
\usepackage{booktabs}       
\usepackage{amsfonts}       
\usepackage{nicefrac}       
\usepackage{microtype}      
\usepackage{xcolor}         
\usepackage{mathtools}
\usepackage{algorithm}
\usepackage{algorithmic}
\usepackage{hyperref}
\usepackage{graphicx}
\usepackage{subfigure}

\newcommand{\indep}{\rotatebox[origin=c]{90}{$\models$}}

\newcommand{\CondProb}[2]{
    P\left( #1 \mid #2 \right)
}
\newcommand{\docal}[2]{
\mathrm{do}(#1 = #2)
}

\title{Bayesian Intervention Optimization for Causal Discovery}

\author{%
Yuxuan Wang$^1$\\
\And
  Mingzhou Liu$^1$\\
  \And
  Xinwei Sun$^2$\thanks{Correspondence to sunxinwei@fudan.edu.cn, wangwei@bigai.ai} \\
  \And
  Wei Wang$^3$\footnotemark[1] \\
  \And
  Yizhou Wang$^1$ \\
}

\date{\newline\newline
    $^1$ School of Computer Science, Peking University \\ 
    $^2$ School of Data Science, Fudan University \\ 
    $^3$ National Key Laboratory of General Artificial Intelligence, BIGAI \\
}

\begin{document}

\maketitle

\input{paras/abstract}
\input{paras/introduction}
\input{paras/preliminary}
\input{paras/method}
\input{paras/experiments}
\input{paras/conclusion}

\bibliography{neurips_2024}
\bibliographystyle{plain}

\input{paras/appendix}

\end{document}

%% file: paras/abstract.tex
\begin{abstract}
Causal discovery is crucial for understanding complex systems and informing decisions. While observational data can uncover causal relationships under certain assumptions, it often falls short, making active interventions necessary. Current methods, such as Bayesian and graph-theoretical approaches, do not prioritize decision-making and often rely on ideal conditions or information gain, which is not directly related to hypothesis testing. We propose a novel Bayesian optimization-based method inspired by Bayes factors that aims to maximize the probability of obtaining decisive and correct evidence. Our approach uses observational data to estimate causal models under different hypotheses, evaluates potential interventions pre-experimentally, and iteratively updates priors to refine interventions. We demonstrate the effectiveness of our method through various experiments. Our contributions provide a robust framework for efficient causal discovery through active interventions, enhancing the practical application of theoretical advancements.
\end{abstract}

%% file: paras/introduction.tex
\section{Introduction}
Causal discovery is fundamental for advancing Artificial General Intelligence (AGI), enabling agents to build comprehensive causal world models essential for understanding complex systems and testing hypotheses \cite{richens2024robust}. While observational data is abundant and easy to collect, it often falls short in uncovering true causal relationships due to assumptions like the Markov property and faithfulness \cite{spirtes2001causation}. Active interventions are necessary to recover structural causal models, but they can be prohibitively costly in real-world scenarios, such as genetic knockout experiments used to discover causal links between genes and biological traits. Therefore, efficient active intervention algorithms that can quickly and reliably uncover causal relationships are urgently needed.

To address these challenges, we propose a method inspired by the Bayes factor \cite{kass1995bayes}, widely used in hypothesis testing across various fields including Bayesian statistics \cite{van2021bayesian}, physics \cite{abdalla2022cosmology}, and neuroscience \cite{keysers2020using}. The Bayes factor provides a comparative measure of evidence for two competing hypotheses based on observed data, making them valuable for hypothesis testing in uncertain and data-limited environments \cite{kass1995bayes}. Our approach aims to maximize the probability of achieving decisive and correct evidence, thus directly supporting hypothesis testing.

Our method leverages observational data to estimate causal models under different hypotheses and computes a pre-experimental evaluation of the potential success of interventions. By identifying the most effective interventions and iteratively updating our priors using Bayesian methods, we refine our interventions to increase the likelihood of obtaining strong and truthful evidence through Bayesian optimization. This method addresses the shortcomings of existing approaches by providing a direct measure for acquiring decisive and correct evidence rather than relying on information gain, which does not necessarily translate to hypothesis-testing confidence.

\paragraph{Related Works}

Existing approaches to active intervention can be broadly categorized into Bayesian and graph-theoretical methods \cite{gamella2020active}. The Bayesian approach, pioneered by Tong \cite{tong2001active} and Murphy \cite{murphy2001active}, typically involves selecting an intervention that maximizes a Bayesian utility function, often using mutual information between the graph and the post-intervention samples. While these methods have advanced to consider multiple simultaneous interventions \cite{tigas2022interventions}, budget constraints \cite{agrawal2019abcd}, and expert knowledge \cite{masegosa2013interactive}, they still use information gain, which is not directly related to hypothesis testing.

Graph-theoretical methods explore selecting interventions to recover the maximum number of edges in the graph \cite{he2008active, hauser2014two}, minimize intervention costs \cite{kocaoglu2017cost, lindgren2018experimental}, and address hidden variables \cite{kocaoglu2017experimental}. However, these methods often assume perfect identification of the Markov equivalence class and infinite interventional data, which are impractical in real-world scenarios. Additionally, current active intervention methods predominantly use interventional data without fully utilizing observational data.

Our approach addresses these limitations by integrating observational data to model intervention distribution and focusing on identifying decisive and correct hypotheses. By leveraging the Bayes factor, which provides a comparative measure of evidence for choosing the null or alternative hypotheses, we guide our interventions more effectively. Additionally, by utilizing Bayesian optimization, our framework offers a more direct and robust approach to causal discovery, ensuring that our interventions are grounded in strong statistical evidence.

Our contributions can be summarized as follows:
\begin{enumerate}
    \item We propose an active intervention algorithm that maximizes the probability of obtaining decisive and correct evidence in causal discovery.
    \item We develop a Monte Carlo estimation-based Bayesian optimization method for estimating and optimizing the probability of identifying decisive and correct evidence.
    \item We demonstrate the effectiveness of our method compared to existing approaches using generalized linear models and diverse noise distributions.
\end{enumerate}

%% file: paras/preliminary.tex
\section{Preliminary}
\label{sec:preliminary}
\paragraph{Notation}
\label{sec:notation}
In our study, we consider a Directed Acyclic Graph (DAG) $G = (V, E)$, where $V = \{1, \dots, d\}$ represents the set of vertices, and each vertex corresponds to a random variable. These variables, denoted as $\mathbf{X}_V = \{X_1, \dots, X_d\}\in \mathcal{X}^d$, are indexed by $V$. Our initial observational dataset $\mathbf{D} = \{\mathbf{x}_V^{(i)}\}_{i = 1}^n$ consists of instances drawn from the probability distribution $p_G(x_1, \dots, x_d)$, representing the joint distribution of these variables.

\paragraph{Structural Causal Models (SCM)}
\label{sec:scm}
The underlying data generative mechanism of a DAG $G$ can be descirbed by a set of structural equations:

\begin{equation} X_i \coloneqq f_i(X_{\mathrm{pa}_G(i)}, \varepsilon_i) \quad \forall i\in V \end{equation}

Here, $f_i$'s are the causal mechanisms that remain invariant under interventions, $\mathrm{pa}_G(i)$ denotes parents of variable $X_i$ in DAG $G$, and $\varepsilon_i$'s are mutually independent exogenous noise variables. These structural equations embody both the conditional distributions of random variables and the potential impact of interventions. While these functions can be nonparametric, we often consider parametric approximations with parameters $\gamma$. 
A typical example of SCM is the Additive Noise Model (ANM), where the structural equation for each variable is a function of its parents plus a random noise independent with the parent variables:

\begin{equation} 
X_i \coloneqq f_i(X_{\mathrm{pa}_G(i)}; \gamma_i) + \varepsilon_i\quad\varepsilon_i \indep X_{\mathrm{pa}_G(i)}\quad
\forall i\in V
\end{equation}

\paragraph{Do-calculus}
Do-calculus \cite{pearl2009causality} is a framework used to determine the effect of interventions on a set of variables within a causal model. Specifically, an intervention, denoted by $\mathrm{do}(X_\mathbf{W} = \mathbf{x}'_\mathbf{W})$, modifies the original distribution by setting the variables $X_\mathbf{W}$ to specific values $\mathbf{x}'_\mathbf{W}$, effectively breaking any causal dependencies from their parents. The resulting interventional distribution is given by the truncated factorization formula:
\begin{equation*}
\label{eq:dodef}
\begin{aligned}
  P(\mathbf{X}_V \mid \mathrm{do}(\mathbf{X}_W = \mathbf{x}'_W) = \prod_{i \in V \setminus W} P(X_i \mid X_{\text{pa}_G(i)}) \mathbb{I}(\mathbf X_W = \mathbf{x}'_W),
\end{aligned}
\end{equation*}
where $\mathbb{I}(\cdot)$ is an indicator function. A perfect intervention on any variable $X_j$ removes all dependencies with its parents, resulting in a modified DAG $G' = (V, E \setminus \{(\text{pa}_G(j), j)\})$. This approach allows us to isolate the causal effect of the intervention on the target variables, providing a clearer understanding of the underlying causal relationships.





\paragraph{Hypothesis testing with Bayes factors}
\label{sec:bayesfactor}
The Bayes factor \cite{kass1995bayes} is an important tool in Bayesian hypothesis testing, providing a comparative measure of the evidence for two competing hypotheses. Given two hypotheses $ \mathbb{H}_0 $ and $ \mathbb{H}_1 $, and observed data $\mathbf D $, the Bayes factor $\text{BF}$ is defined as the ratio of the likelihoods of the data under each hypothesis:

\begin{equation}
    \text{BF}_{01}=\frac{P\left(\mathbf D\mid \mathbb{H}_0\right)}{P\left(\mathbf D\mid \mathbb{H}_1\right)}
    \label{eq:bfdef}
\end{equation}
This ratio assesses the relative support for each hypothesis provided by the data. A value of $\text{BF}_{01} > 1$ suggests stronger evidence in favor of $\mathbb H_0$. This method is particularly valuable in decision-making processes where uncertainty and limited data are prevalent.

\begin{table}
	\centering
	\caption{Classification for the evidence levels of the Bayes factor $\textnormal{BF}_{01}$ (from \cite{Schonbrodt:Wagen:2017} adapted from \cite{Jeffreys:1961}).}
	\label{tab:BF:categ}
	\begin{tabular}{ccc}
		\hline
		Bayes factor && Evidence Level \\
		\hline
		$> 100$ && Extreme evidence for $\mathbb H_0$ \\
		$30 - 100$ && Very strong evidence for $\mathbb H_0$ \\
		$10 - 30$ && Strong evidence for $\mathbb H_0$ \\
		$3 - 10$ && Moderate evidence for $\mathbb H_0$ \\
		$1 - 3$ && Anecdotal evidence for $\mathbb H_0$ \\
		$1$ && No evidence \\
		$1/3$ - 1 && Anecdotal evidence for $\mathbb H_1$ \\
		$1/10$ - 1/3 && Moderate evidence for $\mathbb H_1$ \\
		$1/30$ - 1/10 && Strong evidence for $\mathbb H_1$ \\
		$1/100$ - 1/30 && Very strong evidence for $\mathbb H_1$ \\
		$< 1/100$ && Extreme evidence for $\mathbb H_1$ \\
		\hline
	\end{tabular}
\end{table}

The Bayes factor can be classified into different categories of evidence strength, which helps interpret its value in the context of supporting $\mathbb{H}_0$ or $\mathbb{H}_1$ \cite{de2004statistical}. Table \ref{tab:BF:categ} summarizes these categories. This classification provides a nuanced understanding of how compelling the data is in favor of one hypothesis over the other. To simplify, consider the evidence levels for $\mathbb{H}_0$ as described in the table above, and apply symmetry for evidence levels supporting $\mathbb{H}_1$.

%% file: paras/method.tex
\section{Methodology}
\label{sec:method}
In this section, we introduce the problem formulation first and detail our methodology for uncovering causal relationships using active sampling and Bayesian optimization within the Additive Noise Model framework. We focus on determining whether there is a causal edge between two random variables by formulating this problem as a hypothesis test. Specifically, the conditional distributions after applying do-calculus for the hypotheses $\mathbb{H}_0$ and $\mathbb{H}_1$ are:
\begin{equation}
\begin{aligned}
\label{eq:hypotheses}
\mathbb{H}_0: p(y \mid \docal{X}{x}) = p(y)\quad \mathbb{H}_1: p(y \mid \docal{X}{x}) = p(y \mid x)
\end{aligned}
\end{equation}


When $\mathbb{H}_1$ holds, the structural causal model is simply $X\to Y$. If $\mathbb{H}_0$ holds, we assume $X \not \to Y$, meaning there is no direct causal influence from $X$ to $Y$. Under this hypothesis, we consider several possible relationships between $X$ and $Y$ \textbf{(1)} $X \indep Y$: $X$ and $Y$ are independent. \textbf{(2)} $X \gets Y$: There is a direct causal influence from $Y$ to $X$. \textbf{(3)} $X \gets U \to Y$: There exists an unobservable confounder $U$ that influences both $X$ and $Y$. Among these scenarios, the case where $X \indep Y$ represents a situation of independence, which is relatively straightforward and can be easily tested using standard independence tests. Therefore, we will not focus on this scenario in our subsequent analysis. Instead, we will concentrate on the more complex relationships where $X$ and $Y$ are either causally related through $Y \to X$ or via a common confounder $C$.

\subsection{\texorpdfstring{$P_{DC}$}{PDC} as optimization objective}
\label{sec:loss}
In this work, we define a specific metric, $ P_{DC} $, to evaluate the probability of obtaining decisive and correct evidence after performing a specific intervention $\docal{X}{x}$. Let $\mathbf{D}_\text{int}$ denote the already obtained interventional data. Our $ P_{DC} $ metric is formulated to determine the probability of achieving decisive and correct evidence by combining the new data obtained from the intervention $\docal{X}{x}$ with the existing interventional data $\mathbf{D}_\text{int}$. We define the probabilities that the experiment yields a correct and decisive result under the hypotheses $\mathbb{H}_0$ and $\mathbb{H}_1$ hold as $ P_{DC}^0 $ and $ P_{DC}^1 $, respectively.

\begin{equation}
\begin{aligned}
\label{eq:pdc0def}
P_{DC}^0(k_0, \mathbf{D}_\text{int},\docal{X}{x}) &:= P\left(\text{BF}_{01}(\mathbf{D}_\text{int} \cup \mathbf{D}_\text{new}) > k_0 \mid \mathbf{D}_\text{int}, \docal{X}{x}, \mathbb{H}_0\right) \\
P_{DC}^1(k_1, \mathbf{D}_\text{int}, \docal{X}{x}) &:= P\left(\text{BF}_{01}(\mathbf{D}_\text{int} \cup \mathbf{D}_\text{new}) < k_1 \mid \mathbf{D}_\text{int}, \docal{X}{x}, \mathbb{H}_1\right) \\
\end{aligned}
\end{equation}
Here, $\mathbf{D}_\text{int}$ represents the interventional data already collected, and $\mathbf{D}_\text{new}$ represents the new data obtained from the intervention $\docal{X}{x}$. In the definitions of $P_{DC}^0$ and $P_{DC}^1$, $\mathbf{D}_\text{new}$ is sampled under the assumptions that $\mathbb{H}_0$ and $\mathbb{H}_1$ hold true, respectively. Using theseedefinitions, the overall $ P_{DC} $ metric can be defined as:
\begin{equation}
\begin{aligned}
\label{eq:pdcdef}
P_{DC}(k_0, k_1, \mathbf{D}_\text{int}, \docal{X}{x}) := &P_{DC}^0(k_0, \mathbf{D}_\text{int},\docal{X}{x}) P(\mathbb{H}_0\mid \mathbf{D}_\text{int}) \\
&+ P_{DC}^1(k_1, \mathbf{D}_\text{int},\docal{X}{x}) P(\mathbb{H}_1\mid \mathbf{D}_\text{int})
\end{aligned}
\end{equation}

In this formulation, $P_{DC}^i(k_0, \mathbf{D}_\text{int},\docal{X}{x})$ represents the probability of obtaining decisive and correct evidence after performing this additional intervention $\docal{X}{x}$ and combining it with the existing data and hypothesis $\mathbb{H}_i$. 
This allows us to select the $x$ that maximizes $P_{DC}$, thereby obtaining decisive and correct statistical evidence. Formally, solving the optimization problem:

\begin{equation}
\label{eq:optim}
\max_x P_{DC}(k_0, k_1, \mathbf{D}_\text{int}, \docal{X}{x})
\end{equation}

enables us to identify the intervention that maximizes the probability of obtaining strong and correct evidence. We will describe the detailed algorithm in the following sections.

\paragraph{Comparison with previous works on $P_{DC}$}
Previous definitions of $P_{DC}$ in the field of Bayesian sample size determination \cite{de2004statistical}, such as $P_{DC}(k_0, k_1, n)$, primarily focused on estimating the required sample size $n$ before conducting experiments to ensure a high probability of obtaining decisive and correct evidence. These methods emphasized determining the number of observations needed, rather than specifying any intervention strategies:

\begin{equation*}
P_{DC}(k_0, k_1, n) = P_{DC}^0(\mathbf{y}^n,k_0 \mid \mathbb{H}_0)P(\mathbb{H}_0) + P_{DC}^1(\mathbf{y}^n,k_1 \mid \mathbb{H}_1)P(\mathbb{H}_1)
\end{equation*}

While these methods aimed to provide substantial evidence for the correct hypothesis, they did not account for active intervention in experiments. Instead, they relied on passively observing the environment's output, focusing solely on the sample size required. Although \cite{castelletti2022bayesian} considered active intervention in experiments, they modeled the sampling strategy as a Gaussian distribution without optimizing the intervention itself to expedite causal discovery. 

Our approach, in contrast, optimizes the intervention variable $\docal{X}{x}$ to maximize the probability of obtaining decisive and correct evidence in a single experiment. This shift in perspective allows us to design interventions that achieve the highest possible evidential support, offering a more targeted and efficient approach to hypothesis testing.

Throughout the following of this paper, we will use the notation $P_{DC}(\mathbf{D}_\text{int}, \docal{X}{x})$ to refer to our defined metric $P_{DC}(k_0, k_1, \mathbf{D}_\text{int}, \docal{X}{x})$ when emphasizing the intervention $\docal{X}{x}$ and the existing data $\mathbf{D}_\text{int}$ and $P_{DC}$ for simplification and metric notation.

\subsection{Optimization for the \texorpdfstring{$P_{DC}$}{PDC}}
In this subsection, we discuss our approach to optimizing the $P_{DC}$ metric using Bayesian optimization. We employ Monte Carlo (MC) sampling to estimate $P_{DC}$ and use gradient descent to identify the optimal intervention. Specifically, we detail the process of estimating the distributions $m_0$ under $\mathbb{H}_0$ and $m_1$ under $\mathbb{H}_1$, and subsequently computing the Bayes  ($\text{BF}$) as defined in Eq. (\ref{eq:bfdef}).

\paragraph{Estimation of \texorpdfstring{$P_{DC}$}{PDC}}
To estimate and optimize $P_{DC}$, we propose a Monte Carlo (MC) estimation method that provides a differentiable surrogate estimator as follow. The comprehensive derivation of this method is available in Appendix \ref{sec:apd_pdc}.

\begin{equation}
\begin{aligned}
\label{eq:pdccal}
P_{DC}(\mathbf{D}_\text{int}, \docal{X}{x}) &\approx\frac{1}{N}\sum_{i=1}^N \exp\left(-\frac{1}{\beta}\text{ReLU}(k_0-\text{BF}_{01}(\mathbf{D}_\text{int}\cup \{y_{0i},x\}))\right)P(\mathbb H_0\mid \mathbf{D}_\text{int})\\
&+\frac{1}{N}\sum_{i=1}^N \exp\left(-\frac{1}{\beta}\text{ReLU}(\text{BF}_{01}(\mathbf{D}_\text{int}\cup \{y_{1i},x\})-k_1)\right)P(\mathbb H_1\mid \mathbf{D}_\text{int})
\end{aligned}
\end{equation}
where $y_{0i} \sim m_0(y)$ is one of $N$ samples drawn from the estimated distribution $m_0(\cdot)$ and $y_{1i} \sim m_1(y\mid x)$ is one of $N$ samples drawn from the estimated distribution $m_1(\cdot\mid x)$. $\beta$ is the scale parameter to approximate the indicator function detailed in the Appendix \ref{sec:apd_pdc}.
Similarly, the second term in the equation can be estimated by modifying the indicator function to reflect the corresponding event under $\mathbb{H}_1$ and sampling from the distribution $m_1(y \mid x)$.

\paragraph{Estimation of the interventional distribution}
\label{sec:sceprior}
Based on our introduction in Section \ref{sec:loss}, we need to estimate the interventional distributions under each hypothesis to sample from them and compute the Bayes factor. Although the observational data has not undergone do-calculus, it can be used to estimate the probability density functions under each hypothesis. Specifically:

\begin{itemize}
    \item Under $\mathbb{H}_0$, the hypothesis implies that $p(y) = p(y \mid \docal{X}{x}, \mathbb{H}_0)$, and we estimate $m_0(y) \approx p(y) = p(y \mid \docal{X}{x}, \mathbb{H}_0)$.
    \item Under $\mathbb{H}_1$, the hypothesis implies that $p(y \mid x) = p(y \mid \docal{X}{x}, \mathbb{H}_1)$, and we estimate $m_1(y \mid x) \approx p(y \mid x) = p(y \mid \docal{X}{x}, \mathbb{H}_1)$.
\end{itemize}

When $\mathbb{H}_0$ holds, we can estimate $m_0$ by performing maximum likelihood estimation (MLE) on the observational data $\{y_i\}_{i=1}^N$. Formally, this estimation is:
\begin{equation*}
\max_{\phi} \sum_{i=1}^{N_\text{obs}} \log m_0(y_i)
\end{equation*}
where $\{y_i,x_i\} \in D_\text{obs}$. Since the mixture of normals can approximate any distribution \cite{dalal1983approximating}, we parameterize $m_0$ using a mixture of normals and solve the above problem.

When $\mathbb{H}_1$ holds, we can write down the structural equations \ref{eq:sceh1}, and we have
\begin{equation*}
\begin{aligned}
\label{eq:sceh1}
X = n_X, \quad Y = f(X, \gamma) + n_Y
\end{aligned}
\end{equation*}
\begin{equation*}
p(y \mid \docal{X}{x}, \mathbb{H}_1) = p(y \mid x) = p_Y(y - f(x, \gamma))
\end{equation*}
We similarly parameterize $m_1$ using a mixture of normals to approximate the ground truth interventional distribution as:
\begin{equation*}
m_1(y \mid x) = p_\psi(y - f(x, \theta)) \approx p(y \mid \docal{X}{x}, \mathbb{H}_1)
\end{equation*}
where $\theta$ and $\psi$ are the parameters of the link function and the mixture of normals for $m_1$, respectively. To estimate the interventional distribution under $\mathbb{H}_1$, we solve the following optimization problem:
\begin{equation*}
\begin{aligned}
\max_{\theta, \psi} \log p(X, Y) 
\Leftrightarrow \max_{\theta, \psi} \sum_{i=1}^{N_\text{obs}} \log p(x_i) + \log m_1(y_i \mid x_i) 
\Leftrightarrow \max_{\theta, \psi} \sum_{i=1}^{N_\text{obs}} \log m_1(y_i \mid x_i)
\end{aligned}
\end{equation*}
where $\{y_i, x_i\} \in D_\text{obs}$.
By parameterizing $m_0$ and $m_1$ using mixtures of normals, we effectively approximate the interventional distributions under $\mathbb{H}_0$ and $\mathbb{H}_1$, respectively. This allows us to sample from these estimated distributions and compute the Bayes factor necessary for our analysis.

\paragraph{Estimation of Bayes Factor}
\label{sec:bfcal}
We employ a non-parametric method to estimate the Bayes factor. Given that we have obtained the maximum likelihood estimates (MLE) $\hat{m}_0$ and $\hat{m}_1$ for the densities $m_0$ and $m_1$ respectively, we can compute the likelihoods directly. Using these likelihoods, we calculate the Bayes factor for $N_\text{int}$ interventional data points $\mathbf D_\text{int}={y_i, x_i}_{i=1}^{N_\text{int}}$ as follows:

\begin{equation*}
\begin{aligned}
\text{BF}_{01}(\mathbf{D}_\text{int}\cup\{y,x\}) &= \frac{P(\mathbf{D}_\text{int}\cup\{y,x\} \mid \hat{m}_0, \mathbb{H}_0)}{P(\mathbf{D}_\text{int}\cup\{y,x\} \mid \hat{m}_1, \mathbb{H}_1)} = \prod_{i=1}^{N_\text{int}} \frac{\hat{m}_0(y_i)}{\hat{m}_1(y_i \mid x_i)} \times \frac{\hat{m}_0(y)}{\hat{m}_1(y \mid x)}
\end{aligned}
\end{equation*}

With these estimates, we can optimize $P_{DC}$ through Bayesian optimization. Based on the optimization results, we actively intervene in the environment to obtain interventional samples. Once new samples are collected, we need to update our estimates of $P_{DC}$ refining our calculations accordingly.

\paragraph{Update of the prior of the hypothesis}
As we actively engage with the environment and collect interventional data, we update the posterior for the hypothesis. This is crucial for calculating $P_{DC}$ in Eq. (\ref{eq:pdccal}) and evaluating the posterior of the ground truth given the interventional data. Using Bayes' theorem, the posterior updates are:

\begin{equation}
\begin{aligned}
\label{eq:updateprior}
P\left(\mathbb{H}_0 \mid \mathbf{D}_\text{int}\right) &= \frac{P\left(\mathbf{D}_\text{int} \mid \mathbb{H}_0\right) P(\mathbb{H}_0)}{\sum_{i\in[0,1]} P\left(\mathbf{D}_\text{int} \mid \mathbb{H}_i\right) P(\mathbb{H}_i)}
\quad 
P\left(\mathbb{H}_1 \mid \mathbf{D}_\text{int}\right) &= \frac{P\left(\mathbf{D}_\text{int} \mid \mathbb{H}_1\right) P(\mathbb{H}_1)}{\sum_{i\in[0,1]} P\left(\mathbf{D}_\text{int} \mid \mathbb{H}_i\right) P(\mathbb{H}_i)}
\end{aligned}
\end{equation}

\begin{algorithm}
\caption{Actively intervention for hypothesis testing}
\label{alg:mainalgo}
\begin{algorithmic}[1]
		\STATE {\bfseries Input: }{Observational data $\mathbf D_\text{obs}$; intervention sample size $M$}
		\STATE {\bfseries Output: }{$\text{BF}_{01}(\mathbf{D}_\text{int}),\CondProb{\mathbb{H}_0}{ \mathbf D_\text{int}},\CondProb{\mathbb{H}_1}{\mathbf D_\text{int}}$}
    \STATE $\mathbf D_\text{int}=\emptyset$ 
    \STATE $\CondProb{\mathbb{H}_0}{\mathbf D_\text{int}}=\CondProb{\mathbb{H}_1}{\mathbf D_\text{int}}=\frac{1}{2}$
    \STATE Estimate the interventional distribution for the hypothese $\hat m_0,\hat m_1$ with the observational data $\mathbf{D}_\text{obs}$ as described in \ref{sec:sceprior}
\FOR{$m=1,\dots,M$}
  \STATE Estimate $P_{DC}(\mathbf{D}_\text{int}, \docal{X}{x})$ using Monte Carlo method as described in Eq. (\ref{eq:pdccal})
  \STATE Optimize $x_\text{opt}$ to maximize $P_{DC}(\mathbf{D}_\text{int}, \docal{X}{x_\text{opt}})$
 \STATE Sample $y\mid\docal{X}{x_\text{opt}}\sim p_G(y\mid \docal{X}{x_\text{opt}})$ from the environment with intervention
 \STATE $\mathbf{D}_\text{int}\leftarrow \mathbf{D}_\text{int}\cup \{(y,x_\text{opt})\}$
 \STATE Update $P\left(\mathbb{H}_0\mid
 \mathbf D_\text{int}\right),P\left(\mathbb{H}_1\mid
 \mathbf D_\text{int}\right)$ with Eq. (\ref{eq:updateprior})
 \ENDFOR 
 \end{algorithmic}
\end{algorithm}

Now we can effectively calculate the Probability of Decisive and Correct evidence ($P_{DC}$). Due to the differentiability of the entire computational process, we can directly utilize some gradient-based methods for optimization. This capability allows us to precisely refine our intervention strategies and improve the efficiency and accuracy of our causal discovery process. In summary, we propose a Bayesian optimization-based active intervention algorithm, outlined in Algorithm \ref{alg:mainalgo}. The algorithm can be divided into four main steps:

\begin{enumerate}
    \item \textbf{Data-Dependent Prior Formulation}: Using the observational dataset $\mathbf{D}_\text{obs}$ to inform the creation of priors.
    \item \textbf{Monte Carlo Estimation}: Utilizing the formulated priors in a Monte Carlo sampling process to estimate the Probability of Decisive and Correct evidence ($P_{DC}$) for our Bayes factor analysis.
    \item \textbf{Optimization of Intervention Strategy}: Optimizing the intervention strategy $\docal{X}{x}$ to maximize $P_{DC}$.
    \item \textbf{Active Sampling and Updating Priors}: Actively sampling from the environment using the optimized intervention strategy and integrating the newly collected interventional data ($\mathbf{D}_\text{int}$) to update the priors.
\end{enumerate}

%% file: paras/experiments.tex
\section{Experiment}

In this section, we present a series of experiments designed to evaluate the effectiveness of our proposed causal inference methodology. These experiments are conducted on synthetic datasets that represent a variety of causal structures and noise distributions. We generate 5000 observational data points without any intervention for each experimental setup to form the basis of our analysis.

\paragraph{Compared methods}
We compare our method against the following baselines:

\begin{itemize}
    \item \textbf{Random Sampling}: Interventions are chosen uniformly at random.
    \item \textbf{ABCD \cite{agrawal2019abcd}/CBED \cite{tigas2022interventions} Strategies}: These methods maximize information gain to determine the optimal intervention point. One can refer to Appendix \ref{sec:apd_info} for implementation details.
    \item \textbf{Our Method}: A strategy that maximizes the probability of decisive and correct evidence ($P_{DC}$) to determine the intervention value.
\end{itemize}

\paragraph{Evaluation metrics} We evaluate the performance of each method using the following metrics:
 \begin{itemize}
    \item \textbf{$P_{DC}$}: Probability of Decisive and Correct Evidence. A higher $P_{DC}$ indicates a greater chance that the intervention correctly and decisively determines the causal relationship.
    \item \textbf{$\log \text{BF}_{01}$}: The logarithm of the Bayes factor, representing the likelihood ratio of the interventional data under the null hypothesis to that under the alternative hypothesis. Higher values indicate stronger evidence for the null hypothesis, while lower values support the alternative hypothesis.
    \item \textbf{$P(\mathbb{H}_\text{gt} \mid \mathbf{D}_\text{int})$}: Posterior probability of the ground truth hypothesis given the interventional data. Higher values indicate a better likelihood of correctly identifying the true causal relationship, thus are more favorable.
\end{itemize}

\paragraph{Synthetic data generation}
As described in Section \ref{sec:method}, we investigate two hypotheses regarding whether there exists direct causal relationship from $X$ to $Y$.
Previously, we discussed the triviality of the independence scenario ($X \indep Y$) under $\mathbb{H}_0$. Here, we focus on the remaining possible relationships under $\mathbb{H}_0$: \textbf{(1)} $X \gets Y$: There is a direct causal influence from $Y$ to $X$. \textbf{(2)} $X \gets U \to Y$: There exists an unobservable confounding variable $U$ that influences both $X$ and $Y$.

For $\mathbb{H}_1$ ($X \to Y$), the structural equations are:
\begin{equation*}
\begin{aligned}
X = n_X,\quad Y = f(X, \gamma) + n_Y
\end{aligned}
\end{equation*}
where $f$ is parameterized by $\gamma$. In our experiments, we employ a generalized linear model with an invertible link function, specifically $f = \tanh$. The function is parameterized as $A \cdot \tanh(BX)$, with $A = 2$ and $B = 1$.

When considering $\mathbb{H}_0$ under the scenario $X \gets Y$, the structural equations are:
\begin{equation*}
\begin{aligned}
Y = n_Y,\quad X= f(Y, \gamma) + n_X
\end{aligned}
\end{equation*}
using the same parameterization for $f$ as the ground truth of $\mathbb H_1$.

For the confounding scenario $X \gets U \to Y$, the structural equations become:
\begin{equation*}
\begin{aligned}
U = n_U, \quad Y = f(U, \gamma) + n_Y,\quad X = g(U, \beta) + n_X
\end{aligned}
\end{equation*}
where $f$ and $g$ are parameterized similarly, with $f = A \cdot \tanh(BU)$ and $g = C \cdot \tanh(DU)$ , ensuring the unobservable confounder $U$ influences both $X$ and $Y$. We set $A=2,B=1,C=2,D=1$.
In each ground truth scenario, we generate 5000 observational data points based on the specified structural equations. These observational datasets are then provided to all baseline methods for consistent evaluation and comparison.

To represent a wide range of noise distributions in our experiments, we use a mixture of normals due to its flexibility in approximating various distributions \cite{dalal1983approximating}. For each experimental setup, we randomly generate three distributions to ensure a diverse range of noise characteristics. To prevent the distributions from degenerating into overly simple forms, we fix the position and variance of each component within the distributions while randomizing their weights. The weights for the $n$ components are generated using a combination of uniform and softmax distributions. The mixture model is defined as follows:
\begin{equation*}
    n_Y = \sum_{i=1}^{k} \pi_i \mathcal{N}(\mu_i, \sigma_i^2),\quad \mathbf \pi = \frac{1}{2n}\mathbf 1 + \frac{1}{2} \text{Softmax}(\mathbf{z})
\end{equation*}
where $k$ is the number of components, $\pi_i$ are the weights, $\mu_i$ are the means, and $\sigma_i^2$ are the variances of the components, $\mathbf 1_n$ is a $n$-dimension vector of ones and $\mathbf{z} = (z_1, z_2, \ldots, z_n)$ are $n$ independent samples from a standard normal distribution $\mathcal{N}(0, I_n)$. This approach ensures that the weights are both diverse and appropriately normalized, thus avoiding degeneracy. This method allows us to simulate a wide variety of noise distributions, enhancing the robustness of our experiments. For each experimental setup, the ground truth noise distribution is randomly chosen from the three randomly generated distributions to mitigate the impact of randomness. We also conducted experiments to show that even when the position and variance of each component are randomized, the results and conclusions remain consistent. All experiments are repeated with 10 different random seeds to ensure the robustness and reproducibility of our results.

\begin{figure}[!t]
\vskip 0.2in
\begin{center}
\subfigure{
    \includegraphics[width=0.3\columnwidth]{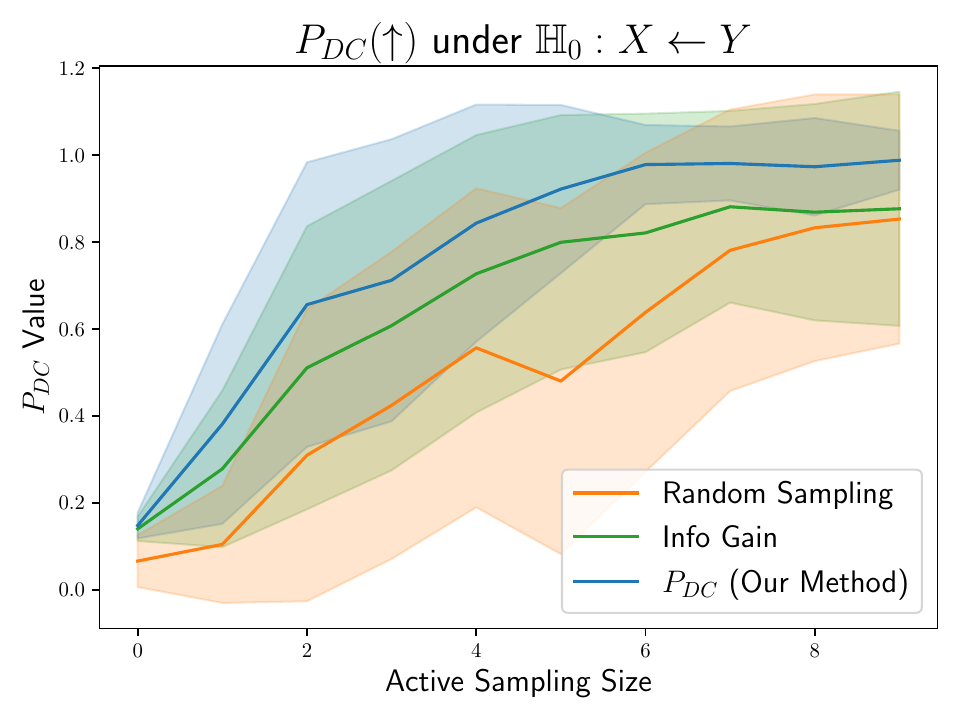}
    \label{fig:pdc_h0_y_to_x}
}
\hfill
\subfigure{
    \includegraphics[width=0.3\columnwidth]{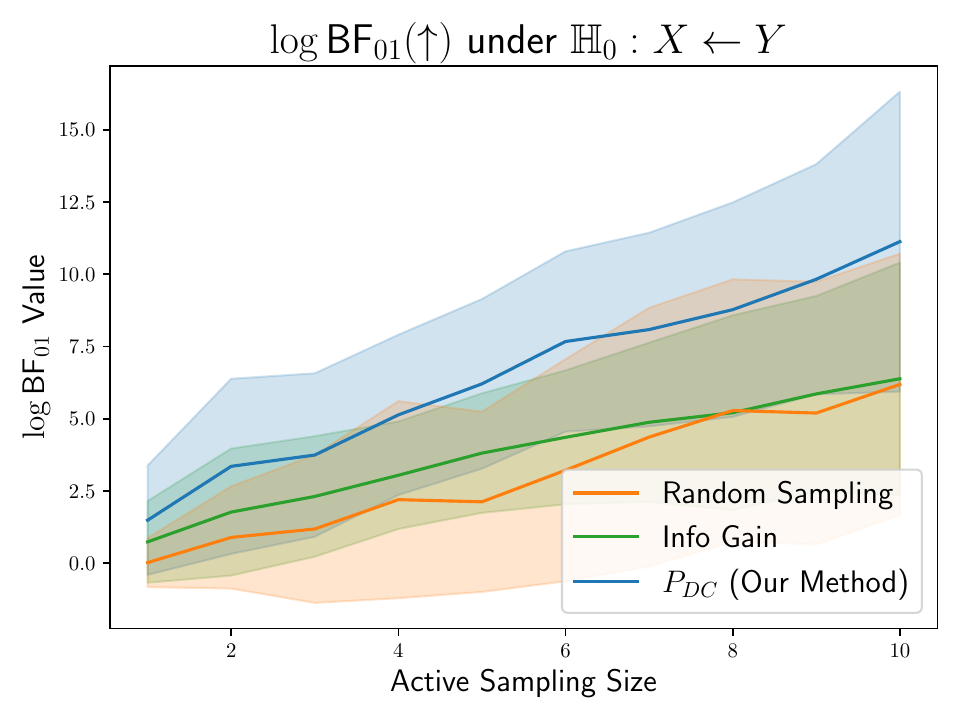}
    \label{fig:log_bf_h0_y_to_x}
}
\hfill
\subfigure{
    \includegraphics[width=0.3\columnwidth]{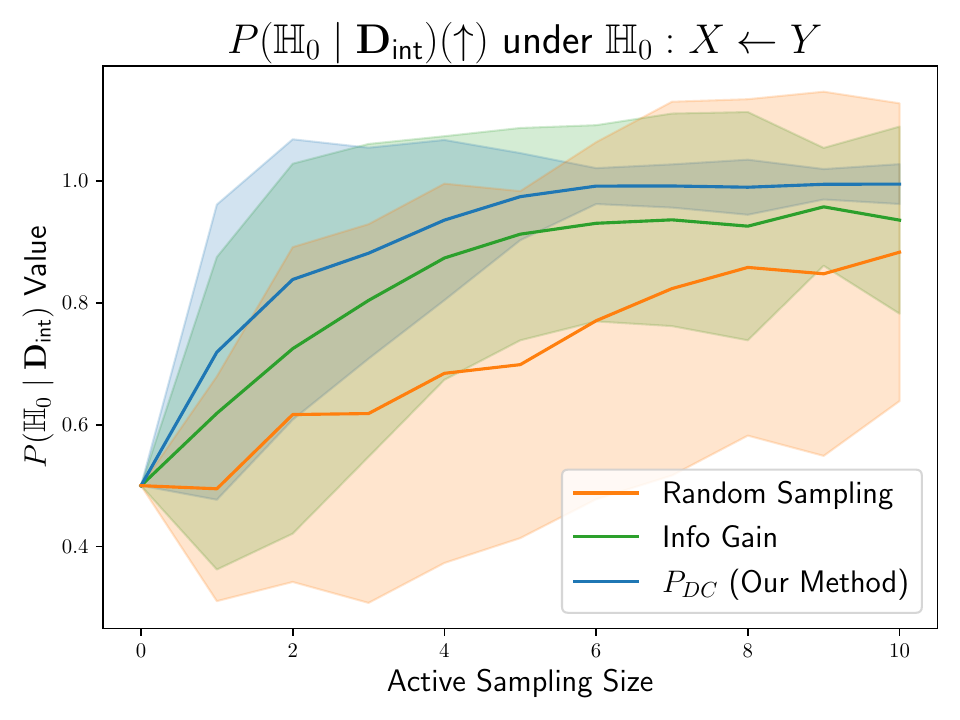}
    \label{fig:ph_gt_h0_y_to_x}
}
\hfill
\subfigure{
    \includegraphics[width=0.3\columnwidth]{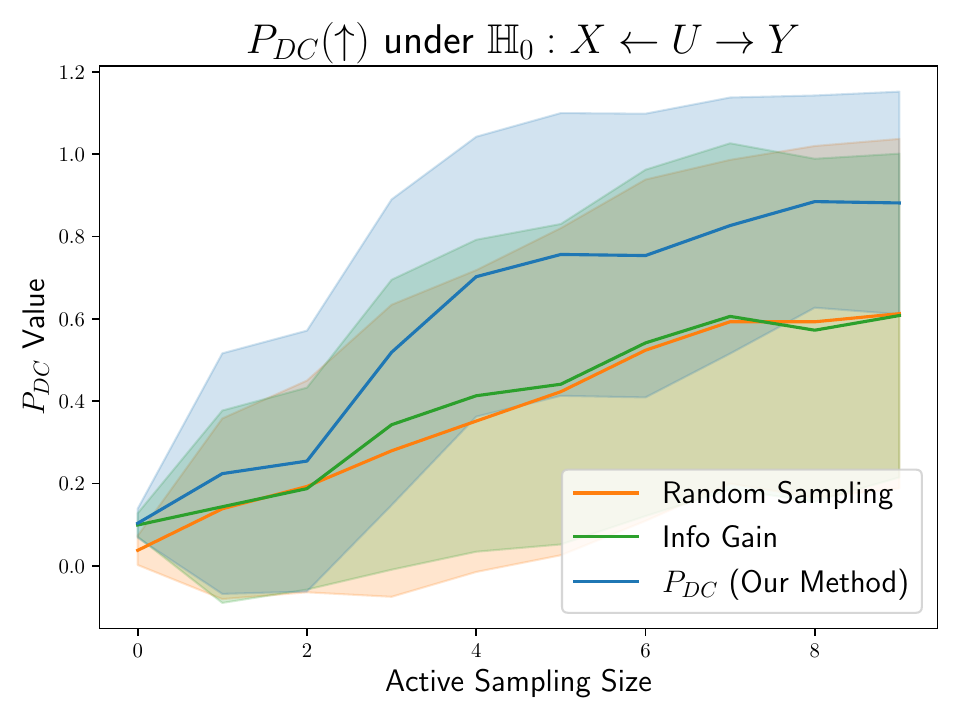}
    \label{fig:pdc_h0_confounder}
}
\hfill
\subfigure{
    \includegraphics[width=0.3\columnwidth]{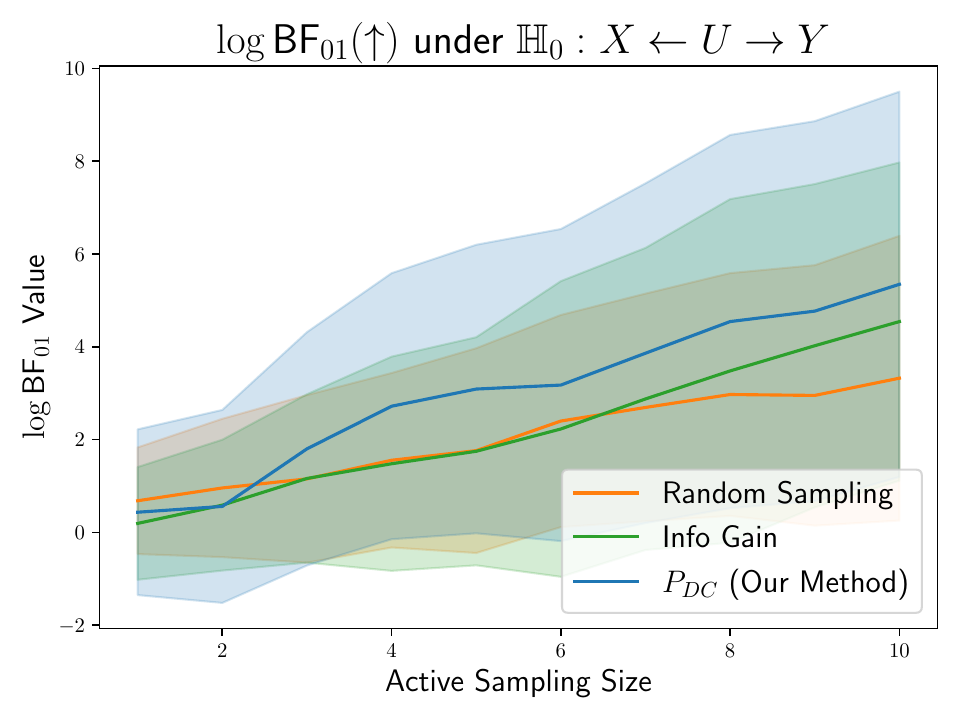}
    \label{fig:log_bf_h0_confounder}
}
\hfill
\subfigure{
    \includegraphics[width=0.3\columnwidth]{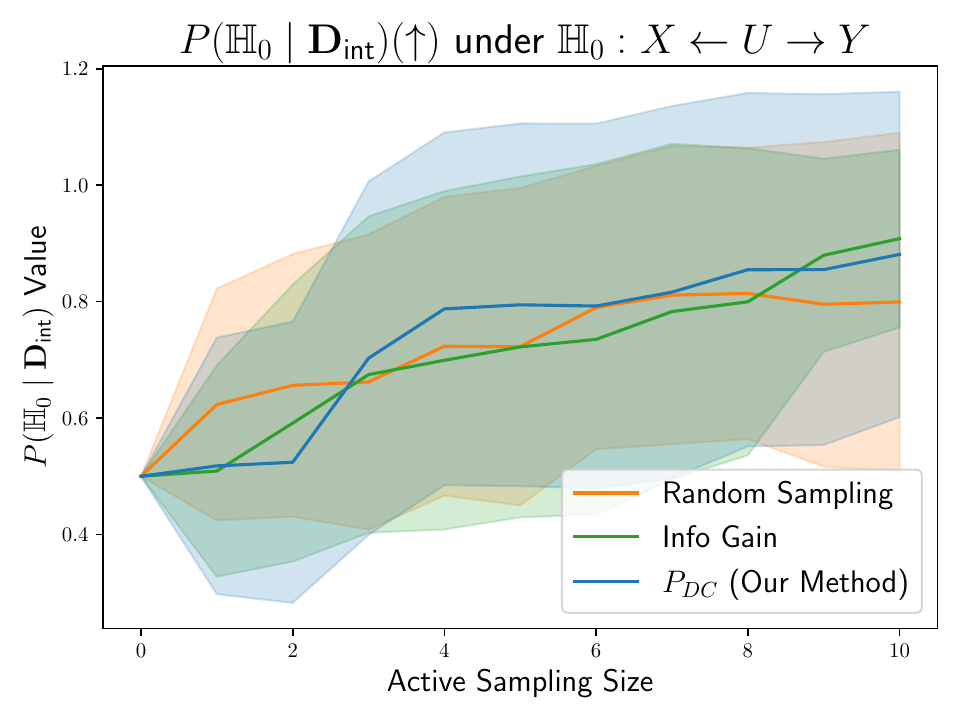}
    \label{fig:ph_gt_h0_confounder}
}
\hfill
\subfigure{
    \includegraphics[width=0.3\columnwidth]{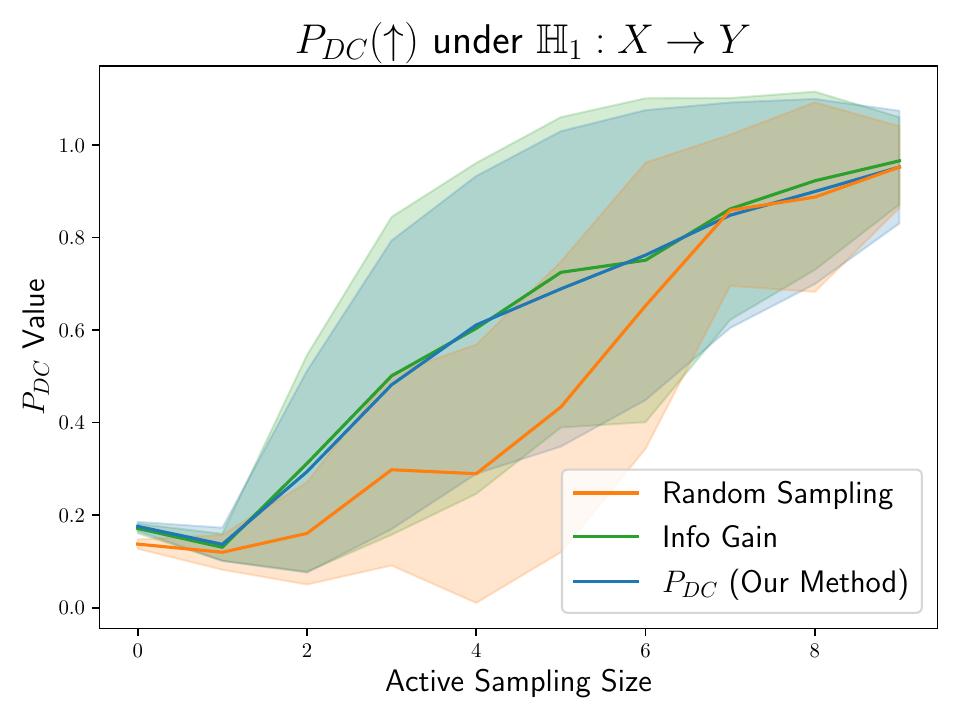}
    \label{fig:pdc_h1_x_to_y}
}
\hfill
\subfigure{
    \includegraphics[width=0.3\columnwidth]{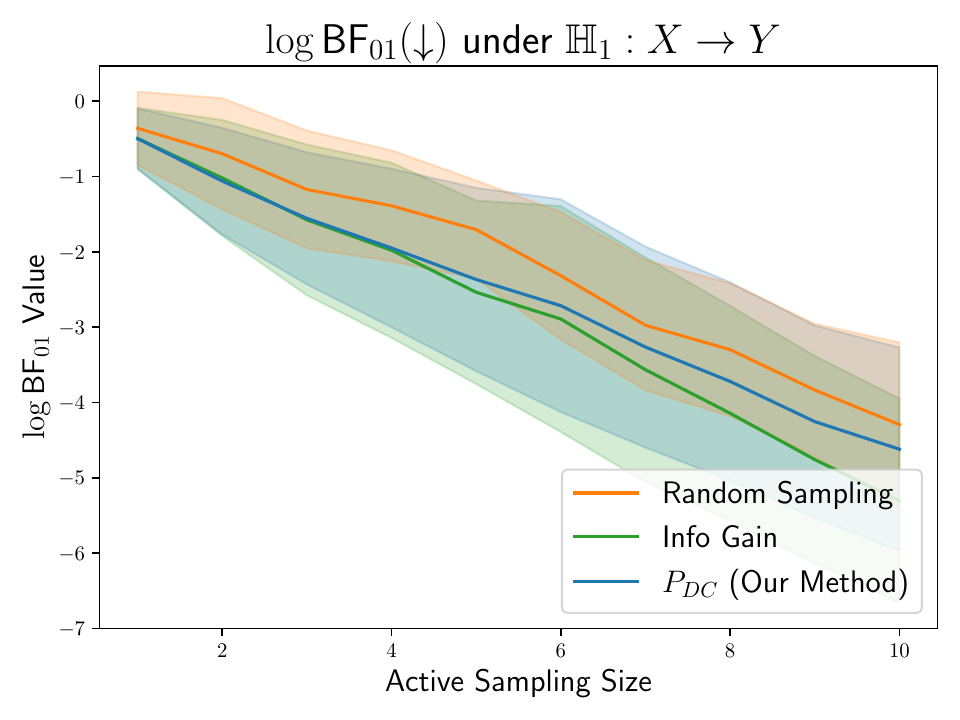}
    \label{fig:log_bf_h1_x_to_y}
}
\hfill
\subfigure{
    \includegraphics[width=0.3\columnwidth]{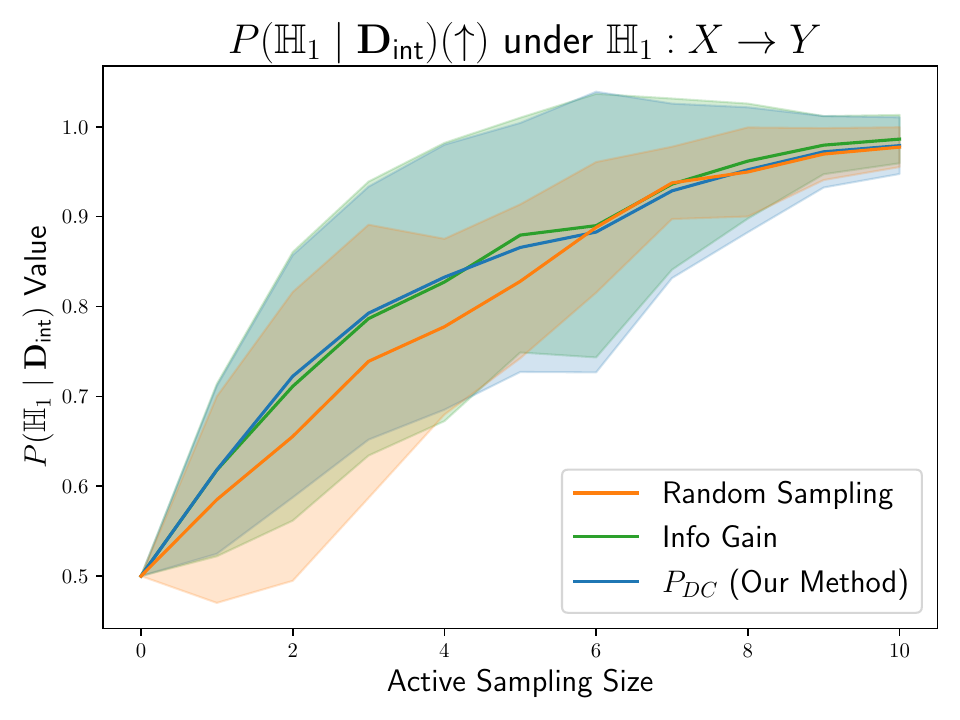}
    \label{fig:ph_gt_h1_x_to_y}
}
\caption{Results under different ground truths with $k_0 = \frac{1}{k_1} = 10$: $P_{DC}$, $\log \text{BF}_{01}$, and $P(\mathbb{H}_{gt} \mid \mathbf{D}_\text{int})$. The first row corresponds to $\mathbb{H}_0$ ($X \gets Y$)). The second row corresponds to $\mathbb{H}_0$ ($X \gets U\to Y$), and the last row corresponds to $\mathbb{H}_1$ ($X \to Y$).}
\label{fig:results_h0_h1}
\end{center}
\vskip -0.2in
\end{figure}

\begin{figure}[!t]
\vskip 0.2in
\begin{center}
\subfigure{
    \includegraphics[width=0.3\columnwidth]{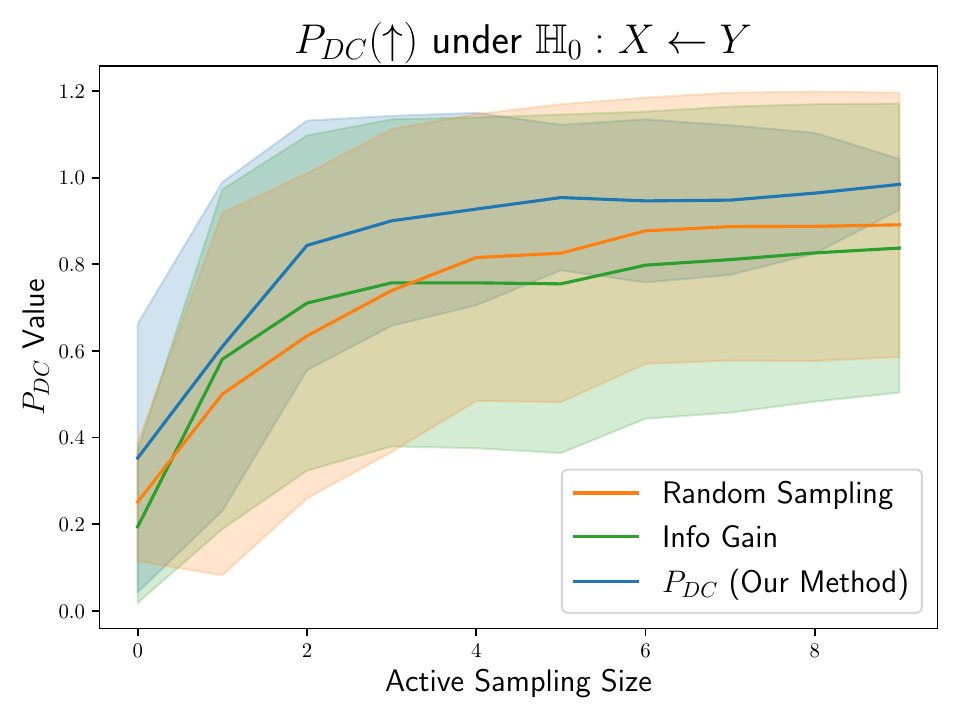}
    \label{fig:all_random_pdc_h0_y_to_x}
}
\hfill
\subfigure{
    \includegraphics[width=0.3\columnwidth]{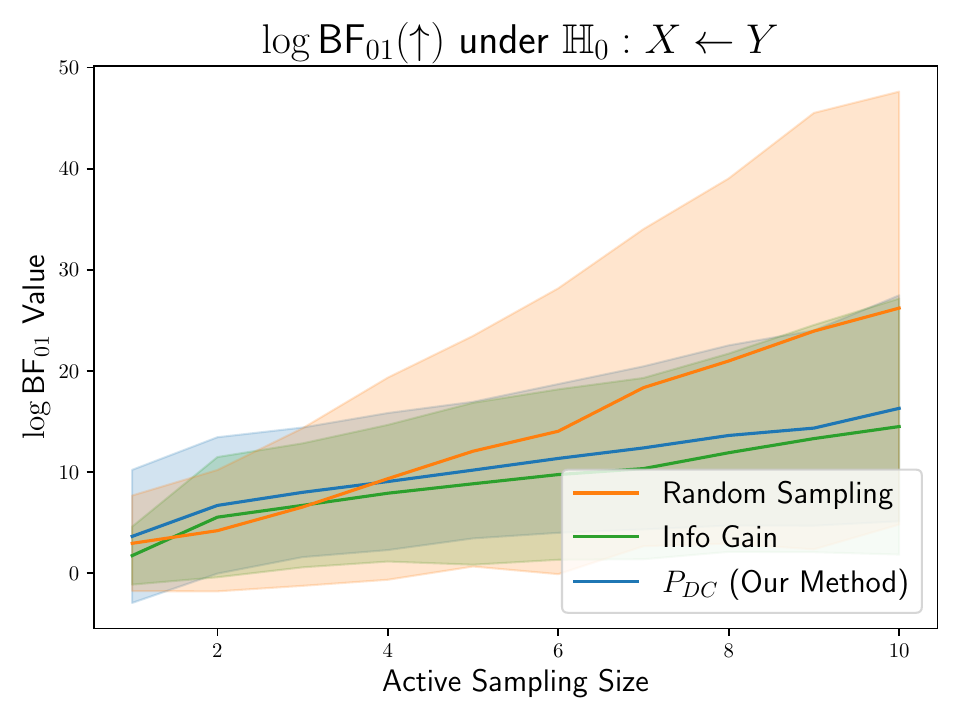}
    \label{fig:all_random_log_bf_h0_y_to_x}
}
\hfill
\subfigure{
    \includegraphics[width=0.3\columnwidth]{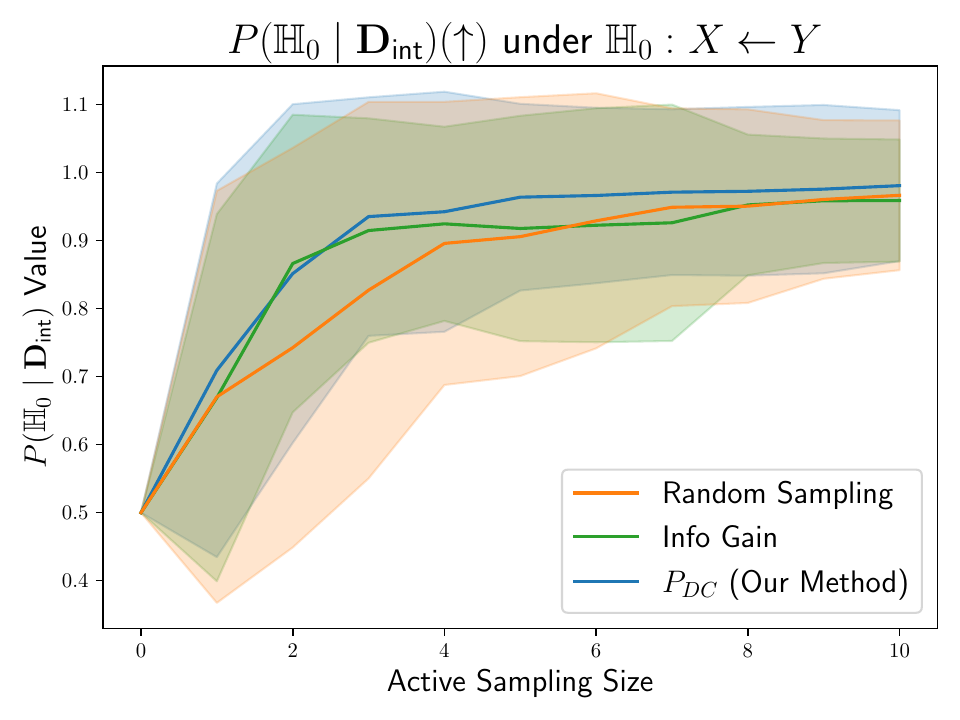}
    \label{fig:all_random_ph_gt_h0_y_to_x}
}
\hfill
\subfigure{
    \includegraphics[width=0.3\columnwidth]{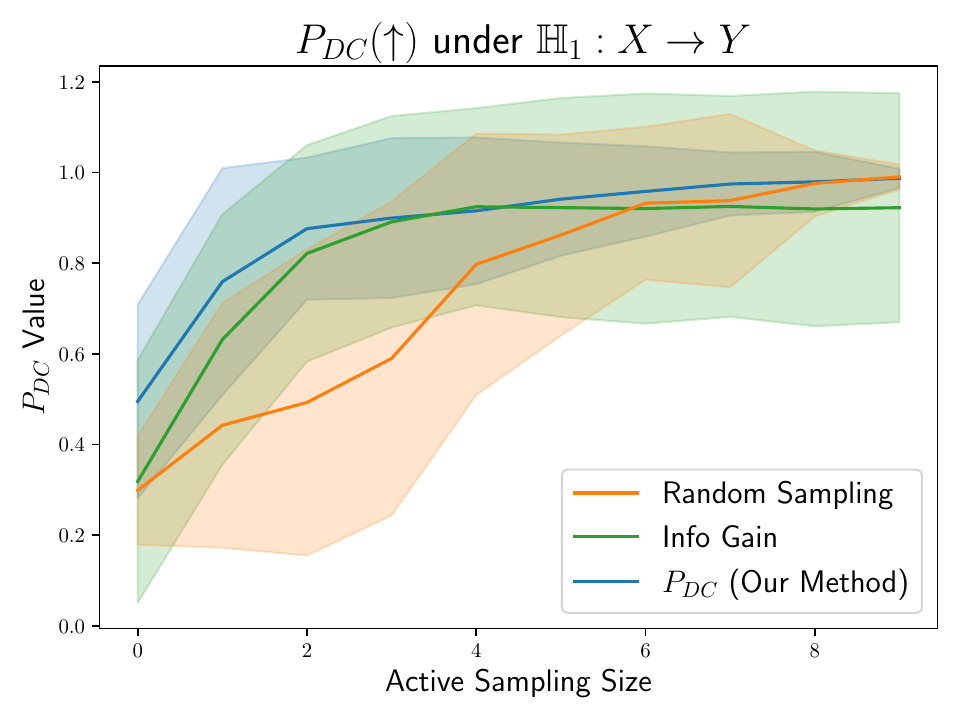}
    \label{fig:all_random_pdc_h1_x_to_y}
}
\hfill
\subfigure{
    \includegraphics[width=0.3\columnwidth]{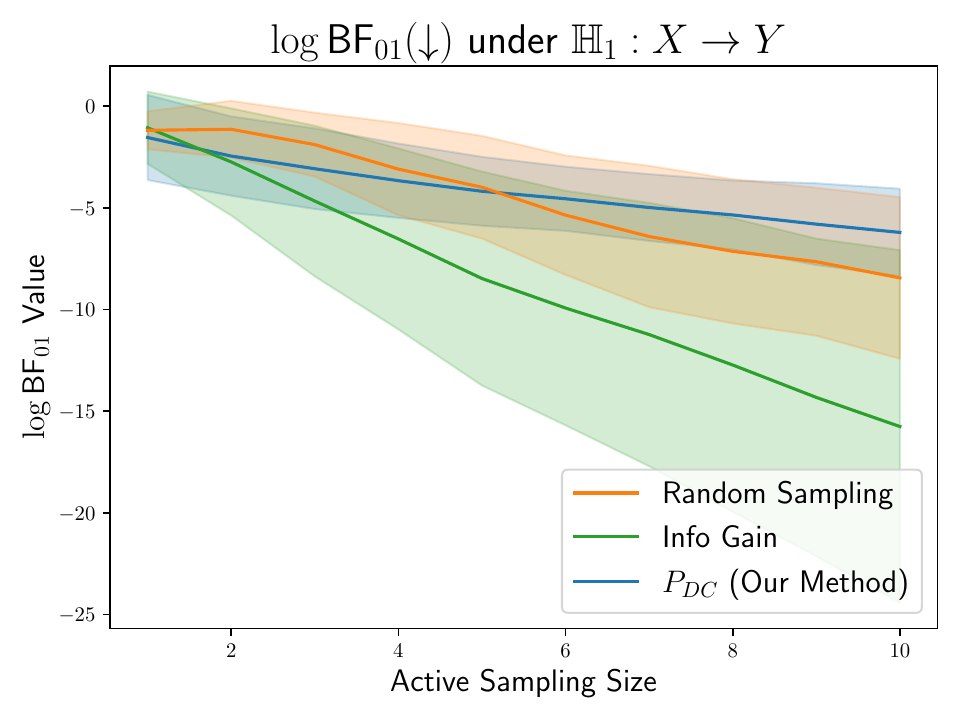}
    \label{fig:all_random_log_bf_h1_x_to_y}
}
\hfill
\subfigure{
    \includegraphics[width=0.3\columnwidth]{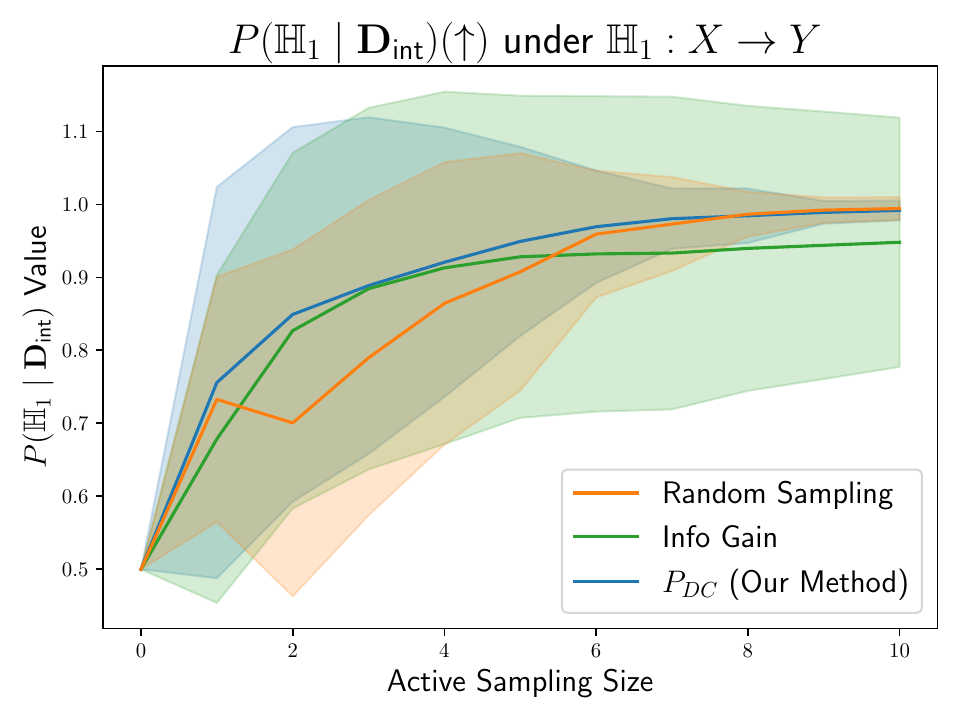}
    \label{fig:all_random_ph_gt_h1_x_to_y}
}
\caption{Results under different ground truths with $k_0 = \frac{1}{k_1} = 10$: $P_{DC}$, $\log \text{BF}_{01}$, and $P(\mathbb{H}_{gt} \mid \mathbf{D}_\text{int})$. But with random mean and random covariance compared to the settings in the Figure \ref{fig:results_h0_h1}.The first row corresponds to $\mathbb{H}_0$ ($X \gets Y$)). The second row corresponds to $\mathbb{H}_0$ ($X \gets U\to Y$), and the last row corresponds to $\mathbb{H}_1$ ($X \to Y$).}
\label{fig:all_random_results_h0_h1}
\end{center}
\vskip -0.2in
\end{figure}

\paragraph{Results}
We delve into the outcomes of our tests corresponding to the earlier described experimental setups. For each hypothesis, we present graphs that illustrate how $P_{DC}$, $\log \text{BF}_{01}$, and $P(\mathbb{H}_{gt} \mid \mathbf{D}_\text{int})$ evolve with the number of samples. Both $P_{DC}$ and $P(\mathbb{H}_{gt} \mid \mathbf{D}_\text{int})$ are higher values indicating better performance, as described in Section \ref{sec:preliminary}. The $\log \text{BF}_{01}$ is evaluated such that higher values are better when $\mathbb{H}_0$ is the ground truth and lower values are better when $\mathbb{H}_1$ is the ground truth. Each graph shows the mean and standard deviation across all experiments and random seeds, with solid lines representing the mean and shaded areas representing the standard deviation.

To demonstrate the effectiveness of our method in obtaining different levels of $P_{DC}$, we present results for various thresholds $k_0$ and $k_1$, specifically $k_0 = \frac{1}{k_1} = 10$. Results for $k_0 = \frac{1}{k_1} = 30$ and $k_0 = \frac{1}{k_1} = 100$ are provided in Appendix \ref{sec:apd_res}. This includes results for different confounders and robustness tests with random means, variances, and weights.

Our results in Figure \ref{fig:results_h0_h1} reveal that when $\mathbb{H}_0$ holds, our algorithm consistently outperforms information gain and random sampling methods in most scenarios and metrics. Specifically, our method demonstrates faster improvement (or decline, depending on whether the metric is maximized or minimized) compared to the baselines. This indicates that our approach is more effective in conducting interventions and achieving decisive and correct evidence. Under $\mathbb{H}_1$, our method performs comparably to existing methods. This could be due to the problem's complexity when $\mathbb{H}_1$ is true, where accumulating a decisive Bayes factor of the alternative hypothesis over the null hypothesis while ensuring correctness is a slower process, leading to a smaller performance gap.

Additionally, our method shows robustness in various randomized settings, including those with random means, variances, and weights as shown in Figure \ref{fig:all_random_results_h0_h1}. While the gap between our method and the baselines decreases in these scenarios, our approach still maintains a competitive edge. This robustness is critical in practical applications where noise and variability are inherent, demonstrating that our method can adapt to different data characteristics and still perform effectively. Our approach's ability to handle diverse conditions while achieving consistent results highlights its practical utility and reliability in real-world causal discovery tasks.

%% file: paras/conclusion.tex
\section{Conclusion and Limitations}

\paragraph{Conclusion}
In this paper, we propose an active sampling and Bayesian optimization-based methodology for hypothesis testing, optimizing the probability of decisive and correct evidence (\(P_{DC}\)). Through extensive experiments on synthetic data, we demonstrate that our method consistently outperforms baseline strategies, including random sampling and information gain-based method in the aspect of the probability of getting the decisive and correct evidence.

\paragraph{Limitations}
A key limitation of our method is the need for prior knowledge of the structural equation's link function when modeling \(\hat{m}_0(y)\) and \(\hat{m}_1(y \mid x)\). This requirement can be restrictive in scenarios where such information is unavailable. Future work could explore approaches that require less specific prior knowledge, potentially employing non-parametric techniques.

%% file: paras/appendix.tex
\newpage
\appendix
\section{Optimization objective and the problems computation derivation}
\subsection{Derivation for equation \ref{eq:pdccal}}
\label{sec:apd_pdc}
Following the definition of $P_{DC}$ in eq \ref{eq:pdcdef}, we have:
\begin{equation*}
\begin{aligned}
\label{eq:apd_pdccal}
P_{DC}&(\mathbf{D}_\text{int}, \docal{X}{x}) = P_{DC}^0(\mathbf{D}_\text{int}, \docal{X}{x})P(\mathbb{H}_0 \mid \mathbf{D}_\text{int}) + P_{DC}^1(\mathbf{D}_\text{int}, \docal{X}{x})P(\mathbb{H}_1 \mid \mathbf{D}_\text{int}) \\
&\text{Expanding $P_{DC}^0$ and $P_{DC}^1$ using their definitions in eq \ref{eq:pdc0def}} \\
&= P\left(\text{BF}_{01}(\mathbf{D}_\text{int} \cup \mathbf{D}_\text{new}) > k_0 \mid \mathbf{D}_\text{int}, \docal{X}{x}, \mathbb{H}_0\right)P(\mathbb{H}_0 \mid \mathbf{D}_\text{int}) \\
&\quad + P\left(\text{BF}_{01}(\mathbf{D}_\text{int} \cup \mathbf{D}_\text{new}) < k_1 \mid \mathbf{D}_\text{int}, \docal{X}{x}, \mathbb{H}_1\right)P(\mathbb{H}_1 \mid \mathbf{D}_\text{int}) \\
&\text{Using the expectation form of the probability of an event $A$, $P(A) = \mathbb{E}[\mathbb{I}(A)]$} \\
&= \mathbb{E}_{y \sim m_0(y)}\left[\mathbb{I}\left(\text{BF}_{01}(\mathbf{D}_\text{int} \cup \{y,x\}) > k_0\right) \mid \mathbf{D}_\text{int}, \docal{X}{x}, \mathbb{H}_0\right]P(\mathbb{H}_0 \mid \mathbf{D}_\text{int}) \\
&\quad + \mathbb{E}_{y \sim m_1(y \mid x)}\left[\mathbb{I}\left(\text{BF}_{01}(\mathbf{D}_\text{int} \cup \{y,x\}) < k_1 \right)\mid \mathbf{D}_\text{int}, \docal{X}{x}, \mathbb{H}_1\right]P(\mathbb{H}_1 \mid \mathbf{D}_\text{int})
\end{aligned}
\end{equation*}

Calculating $P_{DC}$ for general data likelihoods can be extremely challenging due to the complexity and high dimensionality of the data. To address this difficulty, we adopt a Monte Carlo estimation approach combined with optimization techniques. For this approach to be effective, it is crucial that the terms within the Monte Carlo estimator are differentiable. Therefore, we need to handle the indicator functions within the expectations. The indicator function $\mathbb{I}(A)$ is equal to 1 if the condition $A$ is true and 0 otherwise. For our purposes, we transform the indicator function into a Heaviside step function, $H(x)$, which is defined as:

\begin{equation*}
H(x) = 
\begin{cases} 
0 & \text{if } x < 0 \\
1 & \text{if } x \geq 0 
\end{cases}
\end{equation*}

Using this, we can rewrite $\mathbb{I}(A)$ as $H(x - k)$ for some threshold $k$. However, the Heaviside function is not differentiable, and its "derivative" is the Dirac delta function, which is not a function in $L_1$ space but a distribution i.e. genenralized function. To facilitate Monte Carlo estimation and optimization, we need a smooth, differentiable approximation of the Heaviside function.

We approximate the Heaviside function with a smooth function that transitions smoothly around the threshold. One common approach is to use the following exponential smoothing:

\begin{equation*}
\label{eq:halpha}
H_\beta(x) = 
\begin{cases} 
\exp\left(-\frac{x}{\beta}\right) & \text{if } x < 0 \\
1=\exp\left(-\frac{0}{\beta}\right) & \text{if } x \geq 0 
\end{cases}
\end{equation*}

As $\beta$ approaches 0, $H_\beta(x)$ converges to the Heaviside function $H(x)$. By using this smooth approximation, we can maintain differentiability. Noting that $H_\beta$ is in the form of $\exp\left(-\frac{1}{\beta}(\cdot)\right)$, we can rewrite it as $H_\beta(x)=\exp\left(-\frac{1}{\beta}\text{ReLU}(x)\right)$.

With this approximation, we can rewrite the expression. Taking the first term in Eq. \ref{eq:apd_pdccal} as an example:

\begin{equation*}
\begin{aligned}
&\mathbb{E}_{y \sim m_0(y)}\left[\mathbb{I}\left(\text{BF}_{01}(\mathbf{D}_\text{int} \cup \{y,x\}) > k_0 \mid \mathbf{D}_\text{int}, \docal{X}{x}, \mathbb{H}_0\right)\right]P(\mathbb{H}_0 \mid \mathbf{D}_\text{int})\\
&=\mathbb{E}_{y \sim m_0(y)}\left[H\left(\text{BF}_{01}(\mathbf{D}_\text{int} \cup \{y,x\}) - k_0\right) \mid \mathbf{D}_\text{int}, \docal{X}{x}, \mathbb{H}_0\right]P(\mathbb{H}_0 \mid \mathbf{D}_\text{int})\\
&\text{Take a small enought $\beta$}\\
&\approx \mathbb{E}_{y \sim m_0(y)}\left[H_\beta\left(\text{BF}_{01}(\mathbf{D}_\text{int} \cup \{y,x\}) - k_0\right) \mid \mathbf{D}_\text{int}, \docal{X}{x}, \mathbb{H}_0\right]P(\mathbb{H}_0 \mid \mathbf{D}_\text{int})\\
&=\mathbb{E}_{y \sim m_0(y)}\left[\exp\left(-\frac{1}{\beta}\text{ReLU}\left(\text{BF}_{01}(\mathbf{D}_\text{int} \cup \{y,x\}) - k_0\right)\right) \mid \mathbf{D}_\text{int}, \docal{X}{x}, \mathbb{H}_0\right]P(\mathbb{H}_0 \mid \mathbf{D}_\text{int})
\end{aligned}
\end{equation*}

Thus, we replace the indicator function with this smooth approximation in our Monte Carlo estimator. The expectation can then be approximated using Monte Carlo sampling:

\begin{equation*}
\begin{aligned}
P_{DC}(\mathbf{D}_\text{int}, \docal{X}{x}) &\approx\frac{1}{N}\sum_{i=1}^N \exp\left(-\frac{1}{\beta}\text{ReLU}(k_0-\text{BF}_{01}(\mathbf{D}_\text{int}\cup \{y_{0i},x\}))\right)P(\mathbb H_0\mid \mathbf{D}_\text{int})\\
&+\frac{1}{N}\sum_{i=1}^N \exp\left(-\frac{1}{\beta}\text{ReLU}(\text{BF}_{01}(\mathbf{D}_\text{int}\cup \{y_{1i},x\})-k_1)\right)P(\mathbb H_1\mid \mathbf{D}_\text{int})
\end{aligned}
\end{equation*}

In our experiments, we set $N=4096$ and $\beta=0.2$. This choice of parameters ensures a balance between approximation accuracy and computational efficiency.

In this form, the Monte Carlo estimator is differentiable, allowing us to efficiently optimize $P_{DC}$ by selecting the intervention $x$ that maximizes the probability of obtaining decisive and correct evidence.

\subsection{Computation of Information Gain}
\label{sec:apd_info}

In this section, we detail the computation of information gain for causal discovery. Our goal is to estimate the information gain conditioned on the existing interventional data $\mathbf{D}_\text{int}$ when considering the new data $\mathbf{D}_\text{new}$ obtained from the intervention $\docal{X}{x}$. Specifically, we treat the existence of a direct causal link from $X$ to $Y$ (the null hypothesis or the alternative hypothesis) as a random variable, and calculate the mutual information between this variable and the union of the pre-existing interventional samples with the newly obtained data.

When using information gain as a utility function for causal discovery, we aim to maximize the value of the intervention point to achieve the highest possible information gain. According to the conditional information equation \cite{cover1999elements}, we have:

\begin{equation*}
    \begin{aligned}
    \label{eq:condinfo}
        I(\mathbf{D}_\text{new} ; H \mid \mathbf{D}_\text{int}, \docal{X}{x}) = I(\mathbf{D}_\text{int} \cup \mathbf{D}_\text{new} ; H \mid \docal{X}{x}) - I(\mathbf{D}_\text{int} ; H \mid \docal{X}{x})
    \end{aligned}
\end{equation*}

 Let $\mathbf{D} = \mathbf{D}_\text{int} \cup \mathbf{D}_\text{new}$. Note that $\mathbf{D}_\text{int}$ is already a constant and does not contain any randomness, thus $\mathbf{D} \mid \docal{X}{x}, H \sim p(y \mid x, H)$. The second term in Eq. (\ref{eq:condinfo}) is a constant since $\mathbf{D}_\text{int}$ is specified, and we can therefore neglect it.

\begin{equation*}
    \begin{aligned}
    &I(\mathbf{D}_\text{int} \cup \mathbf{D}_\text{new} ; H \mid \docal{X}{x})\\
        &=\mathbb E_{\mathbf D\sim p(y\mid H,\docal{X}{x}), H\sim P(H)}\log\frac{P(\mathbf D,H\mid\docal{X}{x})}{P(\mathbf D\mid \docal{X}{x})P(H\mid \docal{X}{x})}\\
&=\mathbb E_{\mathbf D\sim p(y\mid H,\docal{X}{x}), H\sim P(H)}\log\frac{P(\mathbf D\mid H,\mathbf D_\text{int},\docal{X}{x})}{P(\mathbf D\mid \docal{X}{x})}\\
&=\mathbb E_{\mathbf D\sim p(y\mid H,\docal{X}{x}), H\sim P(H)}\log\frac{P(\mathbf D\mid H,\docal{X}{x})}{\sum_{j\in \{0,1\}}P(\mathbf D\mid H,\docal{X}{x})P(H=\mathbb H_j\mid \docal{X}{x})}\\
&=\mathbb E_{\mathbf D\sim p(y\mid H,\docal{X}{x}), H\sim P(H)}\log\frac{P(\mathbf D\mid H,\docal{X}{x})}{\sum_{j\in \{0,1\}}P(\mathbf D\mid H,\docal{X}{x})P(H=\mathbb H_j)}\\
&=\int p(y\mid  H=\mathbb H_0, \docal{X}{x})P(H=\mathbb H_0)\times \\
&\log\frac{P(\mathbf D\mid H=\mathbb H_0,\docal{X}{x})}{P(\mathbf D\mid H=\mathbb H_0,\docal{X}{x})P(H=\mathbb H_0)+P(\mathbf D\mid H=\mathbb H_1,\docal{X}{x})P(H=\mathbb H_1)}\mathrm dy\\
&+\int p(y\mid  H=\mathbb H_1, \docal{X}{x})P(H=\mathbb H_1)\times \\
&\log\frac{P(\mathbf D\mid H=\mathbb H_1,\docal{X}{x})}{P(\mathbf D\mid H=\mathbb H_0,\docal{X}{x})P(H=\mathbb H_0)+P(\mathbf D\mid H=\mathbb H_1,\docal{X}{x})P(H=\mathbb H_1)}\mathrm dy\\
&=\int p(y\mid  H=\mathbb H_0, \docal{X}{x})P(H=\mathbb H_0)\log\frac{\text{BF}_{01}(\mathbf D)}{\text{BF}_{01}(\mathbf D)P(H=\mathbb H_0)+P(H=\mathbb H_1)}\mathrm dy\\
&+\int p(y\mid  H=\mathbb H_1, \docal{X}{x})P(H=\mathbb H_1)\log\frac{1}{\text{BF}_{01}(\mathbf D)P(H=\mathbb H_0)+P(H=\mathbb H_1)}\mathrm dy\\
    \end{aligned}
\end{equation*}

Given this setup, we recognize that this formulation can be expressed in terms of the Bayes factor. Therefore, we can leverage the previous calculation methods for Bayes factors (Section \ref{sec:bfcal}) and use a Monte Carlo approach similar to that used for estimating and optimizing $P_{DC}$ to estimate and optimize this objective as well.

\section{Experiment results for \texorpdfstring{$k_0=\frac{1}{k_1}=30$}{k0=1/k1=30} and \texorpdfstring{$k_0=\frac{1}{k_1}=100$}{k0=1/k1=100}}
\label{sec:apd_res}
In this section, we present detailed results for the experiment setting. The following results complement the findings discussed in the main text and provide further insights into the performance of our method across varying conditions.

We include results for different confounders and robustness tests with random means, variances, and weights. The method for generating the weights is consistent with the approach described in the main text, where weights are normalized using a combination of uniform and softmax distributions.

For the robustness tests, The means ($\mu_i$) of the components are sampled from a uniform distribution in the range \([-4, 4]\). The variances ($\sigma_i^2$) are sampled from a standard chi-squared distribution with 3 degrees of freedom.

These additional experiments ensure that our method's effectiveness and conclusions remain consistent even under varying noise characteristics and distribution parameters. All experiments are repeated with 10 different random seeds to ensure the robustness and reproducibility of our results.

\begin{figure}[ht]
\vskip 0.2in
\begin{center}
\subfigure{
    \includegraphics[width=0.3\columnwidth]{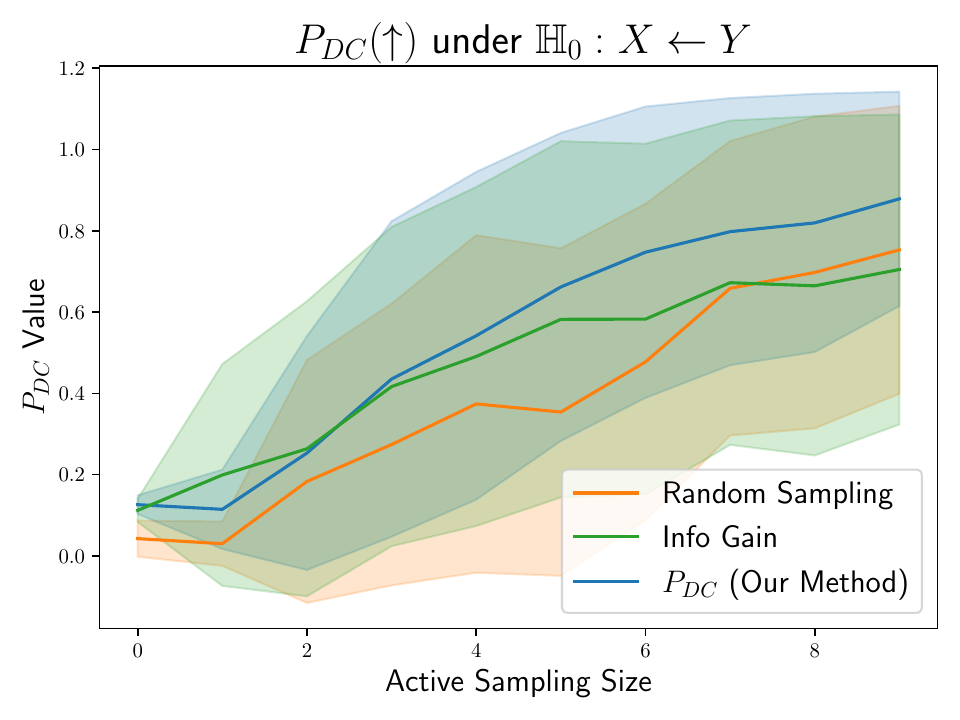}
    \label{fig:k030pdc_h0_y_to_x}
}
\hfill
\subfigure{
    \includegraphics[width=0.3\columnwidth]{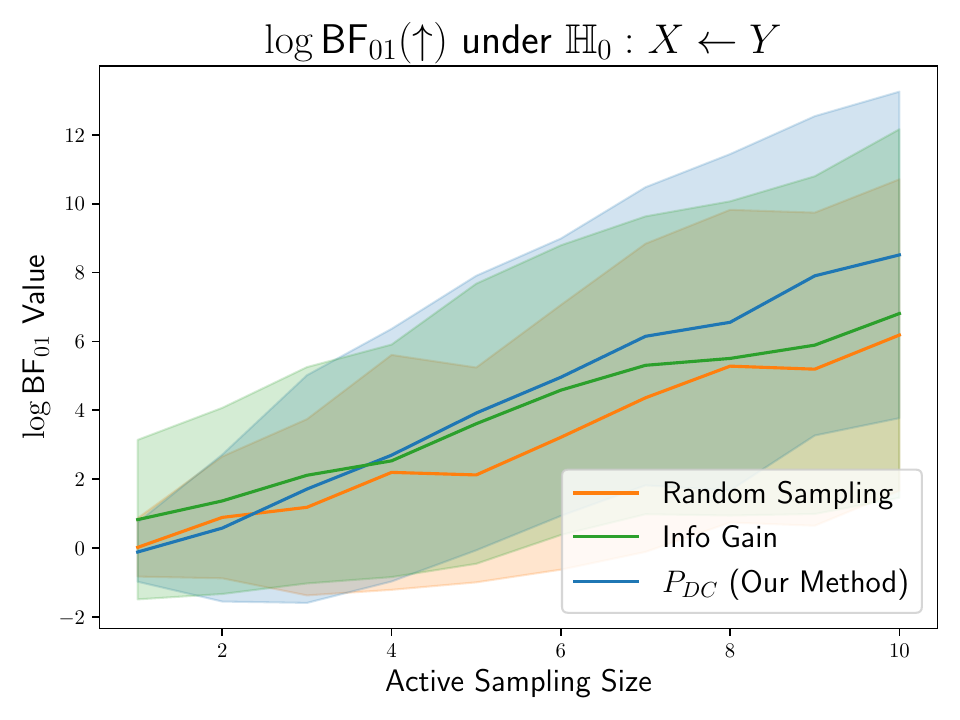}
    \label{fig:k030log_bf_h0_y_to_x}
}
\hfill
\subfigure{
    \includegraphics[width=0.3\columnwidth]{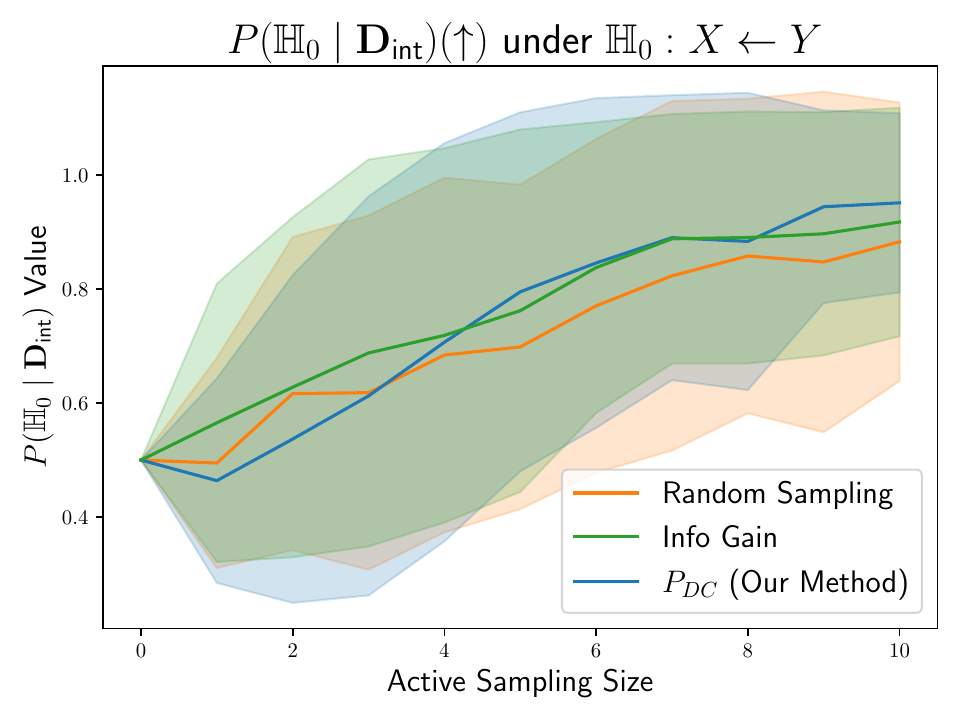}
    \label{fig:k030ph_gt_h0_y_to_x}
}
\hfill
\subfigure{
    \includegraphics[width=0.3\columnwidth]{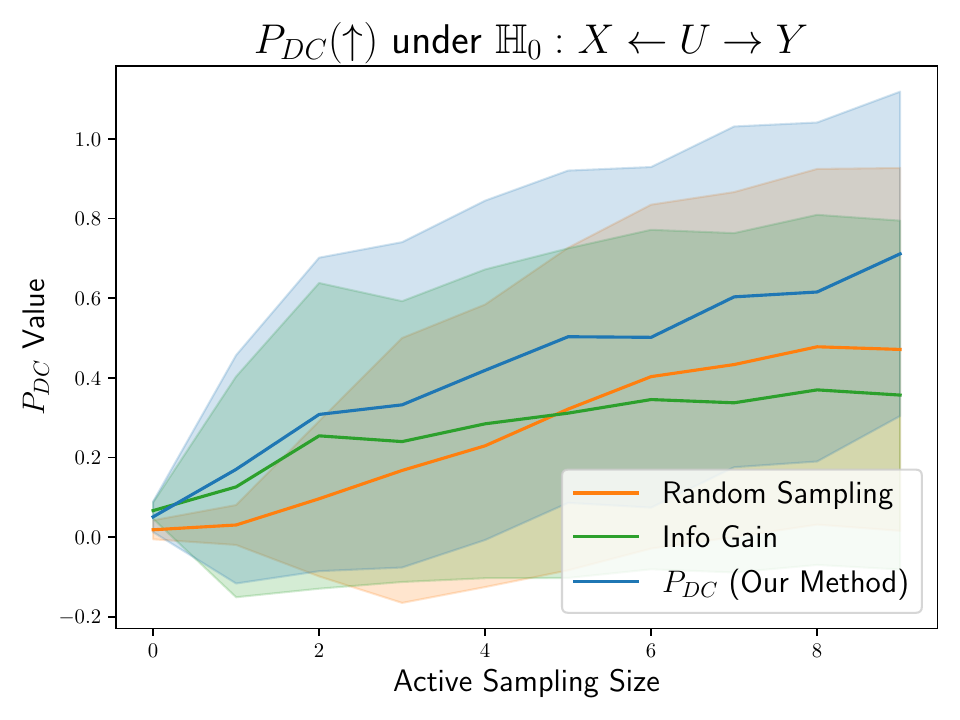}
    \label{fig:k030pdc_h0_confounder}
}
\hfill
\subfigure{
    \includegraphics[width=0.3\columnwidth]{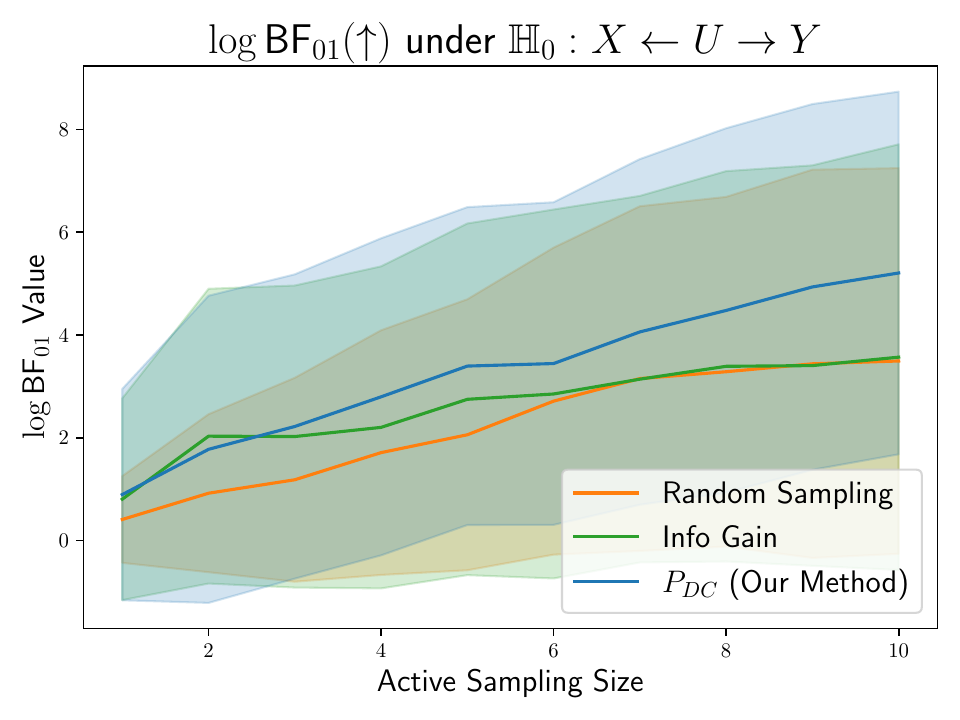}
    \label{fig:k030log_bf_h0_confounder}
}
\hfill
\subfigure{
    \includegraphics[width=0.3\columnwidth]{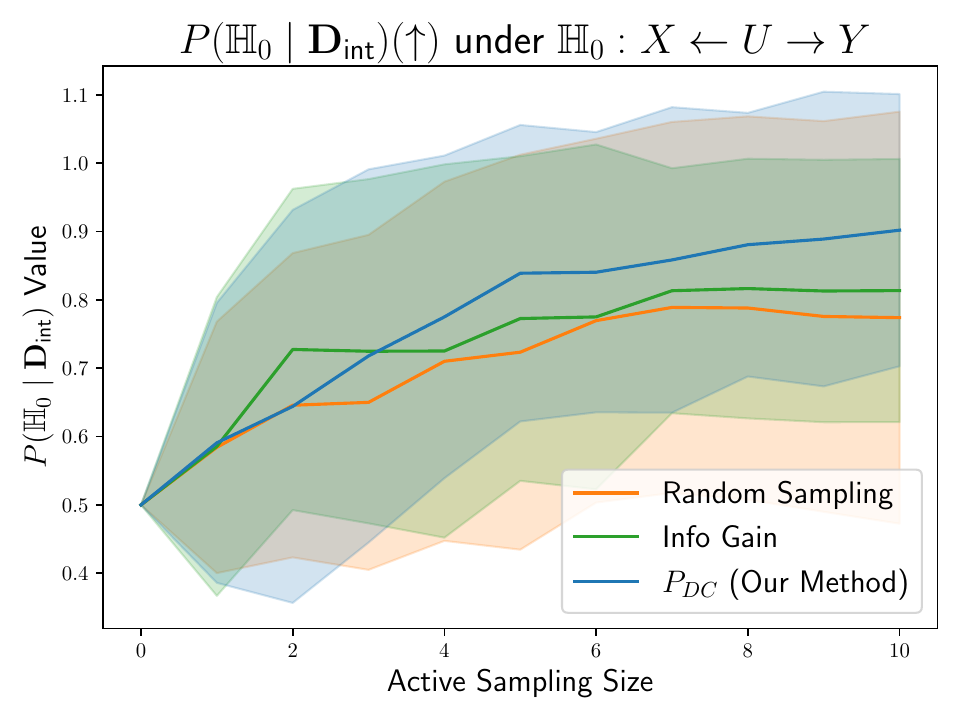}
    \label{fig:k030ph_gt_h0_confounder}
}
\hfill
\subfigure{
    \includegraphics[width=0.3\columnwidth]{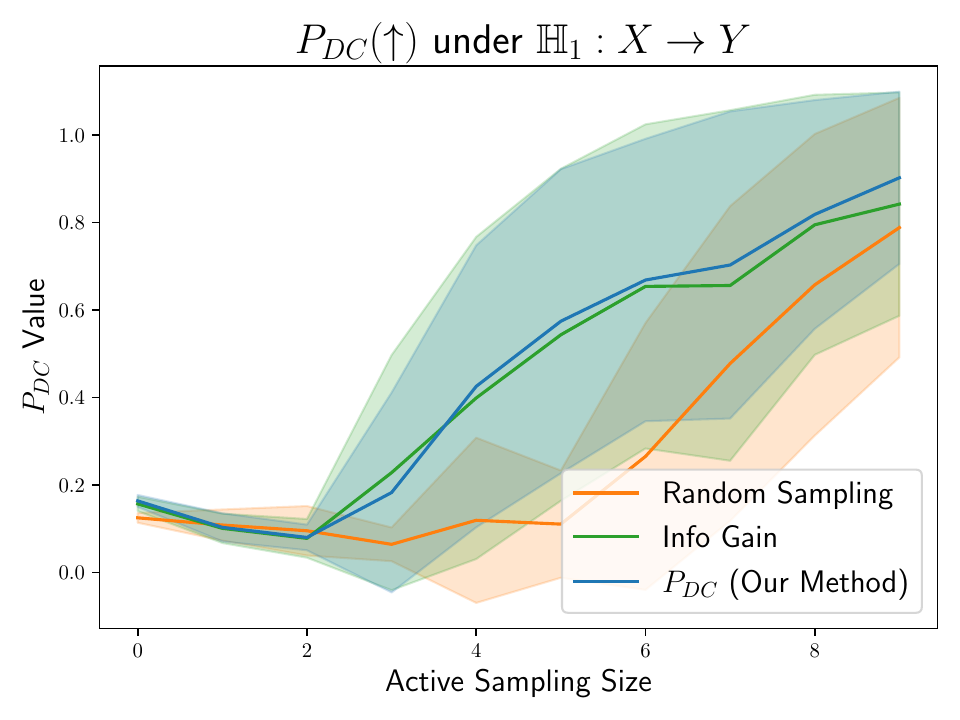}
    \label{fig:k030pdc_h1_x_to_y}
}
\hfill
\subfigure{
    \includegraphics[width=0.3\columnwidth]{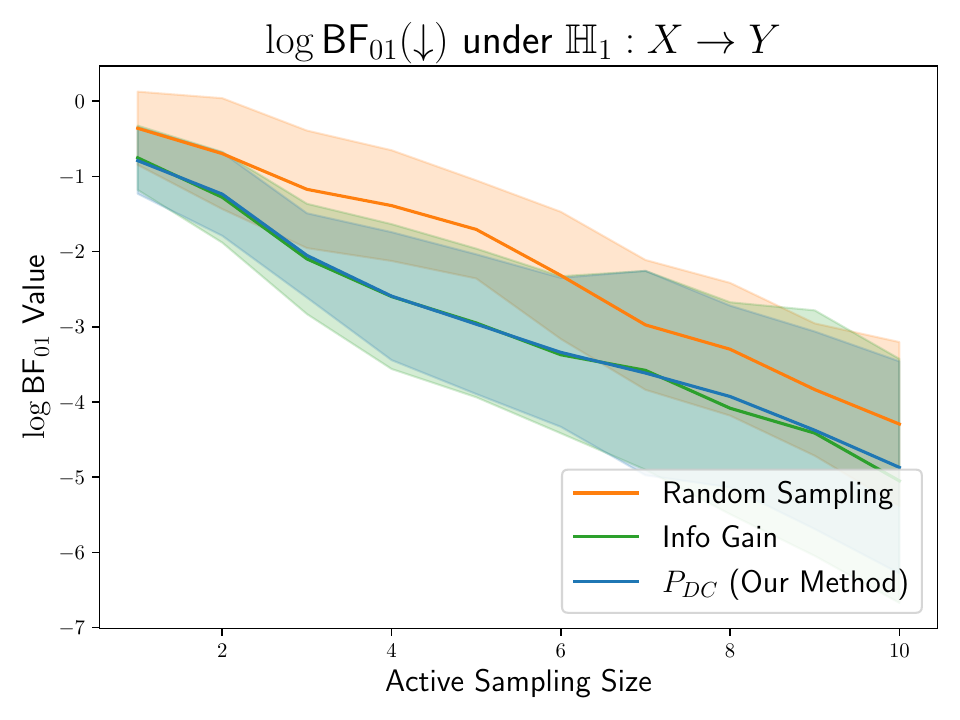}
    \label{fig:k030log_bf_h1_x_to_y}
}
\hfill
\subfigure{
    \includegraphics[width=0.3\columnwidth]{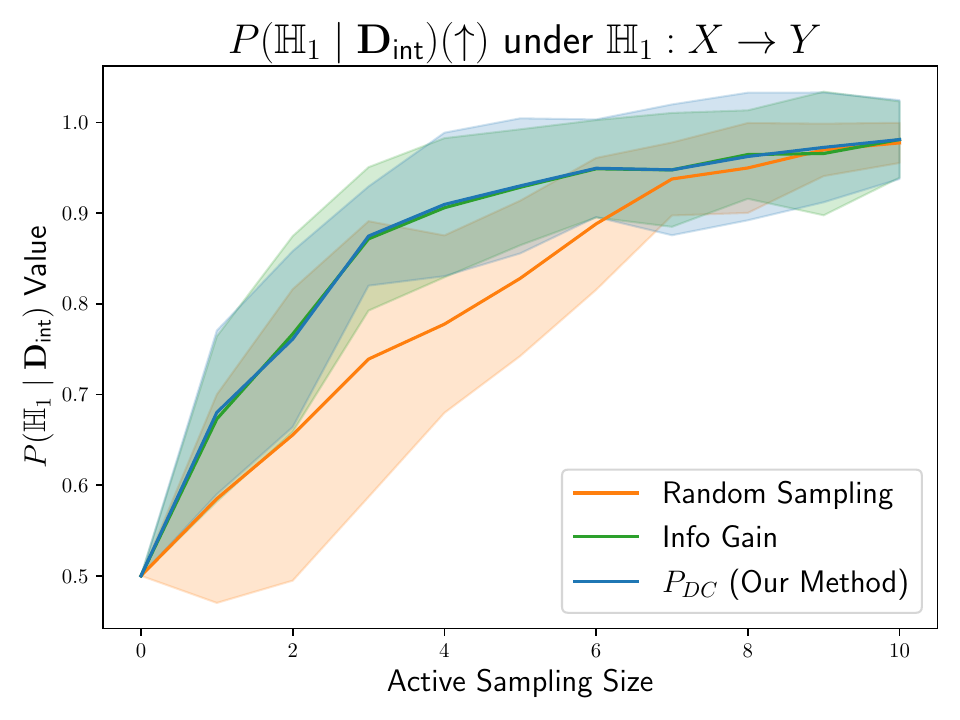}
    \label{fig:k030ph_gt_h1_x_to_y}
}
\caption{Results under different ground truths with $k_0 = \frac{1}{k_1} = 10$: $P_{DC}$, $\log \text{BF}_{01}$, and $P(\mathbb{H}_{gt} \mid \mathbf{D}_\text{int})$. The first row corresponds to $\mathbb{H}_0$ ($X \gets Y$)). The second row corresponds to $\mathbb{H}_0$ ($X \gets U\to Y$), and the last row corresponds to $\mathbb{H}_1$ ($X \to Y$).}
\label{fig:k030results_h0_h1}
\end{center}
\vskip -0.2in
\end{figure}

\begin{figure}[ht]
\vskip 0.2in
\begin{center}
\subfigure{
    \includegraphics[width=0.3\columnwidth]{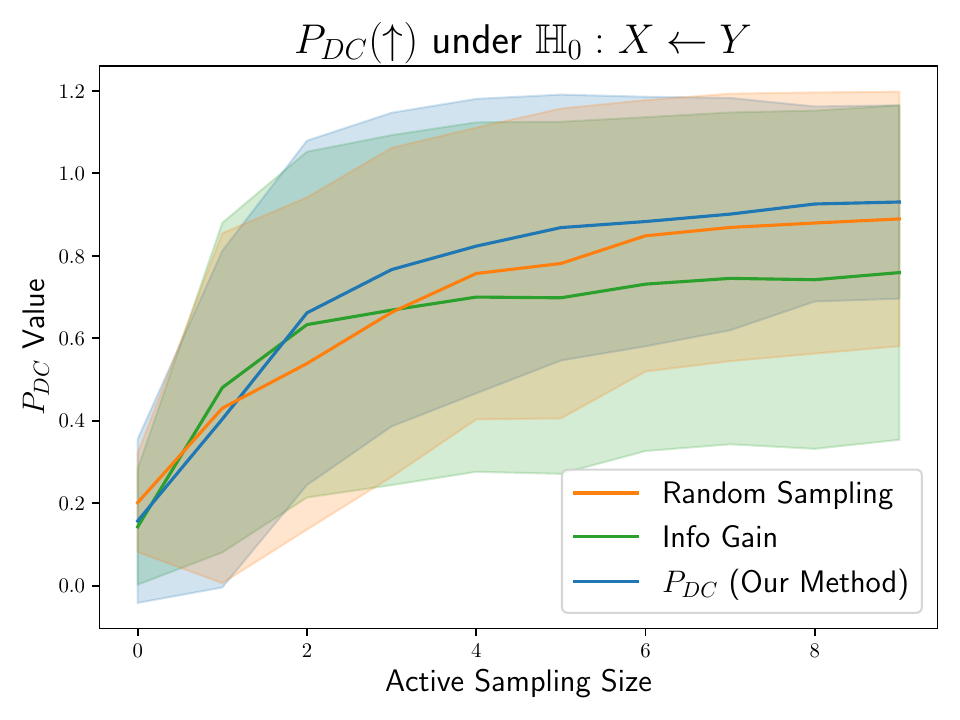}
    \label{fig:all_random_k030pdc_h0_y_to_x}
}
\hfill
\subfigure{
    \includegraphics[width=0.3\columnwidth]{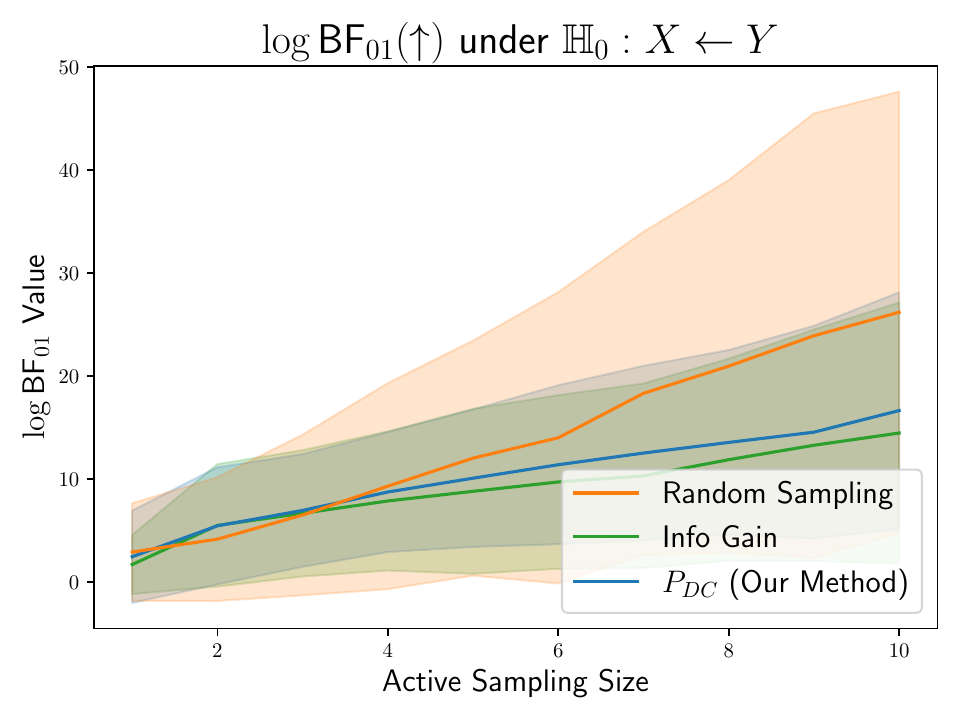}
    \label{fig:all_random_k030log_bf_h0_y_to_x}
}
\hfill
\subfigure{
    \includegraphics[width=0.3\columnwidth]{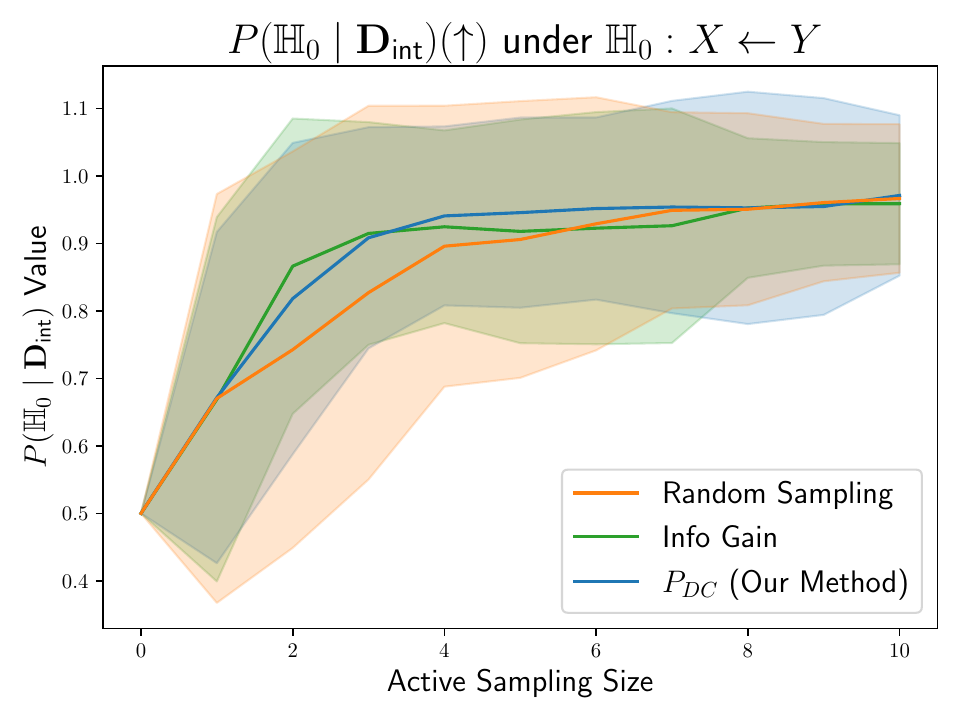}
    \label{fig:all_random_k030ph_gt_h0_y_to_x}
}
\hfill
\subfigure{
    \includegraphics[width=0.3\columnwidth]{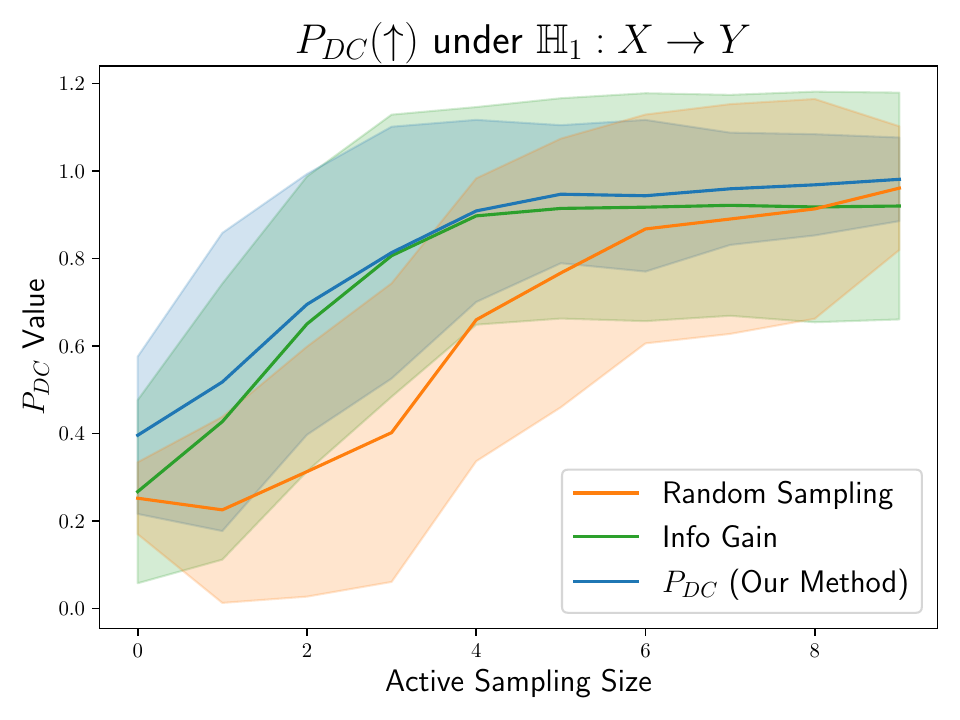}
    \label{fig:all_random_k030pdc_h1_x_to_y}
}
\hfill
\subfigure{
    \includegraphics[width=0.3\columnwidth]{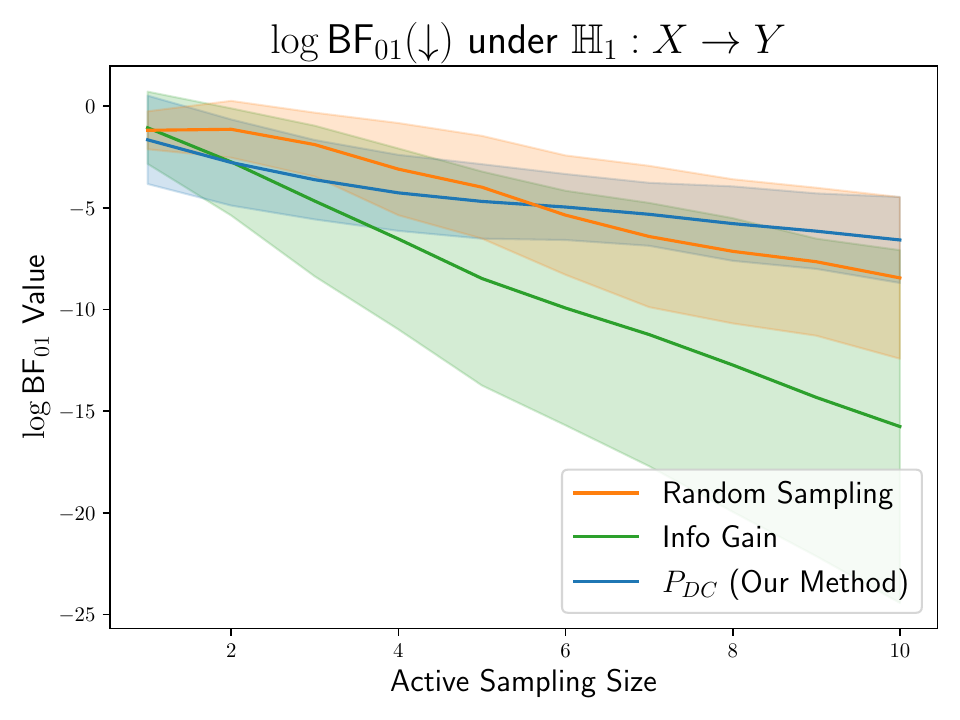}
    \label{fig:all_random_k030log_bf_h1_x_to_y}
}
\hfill
\subfigure{
    \includegraphics[width=0.3\columnwidth]{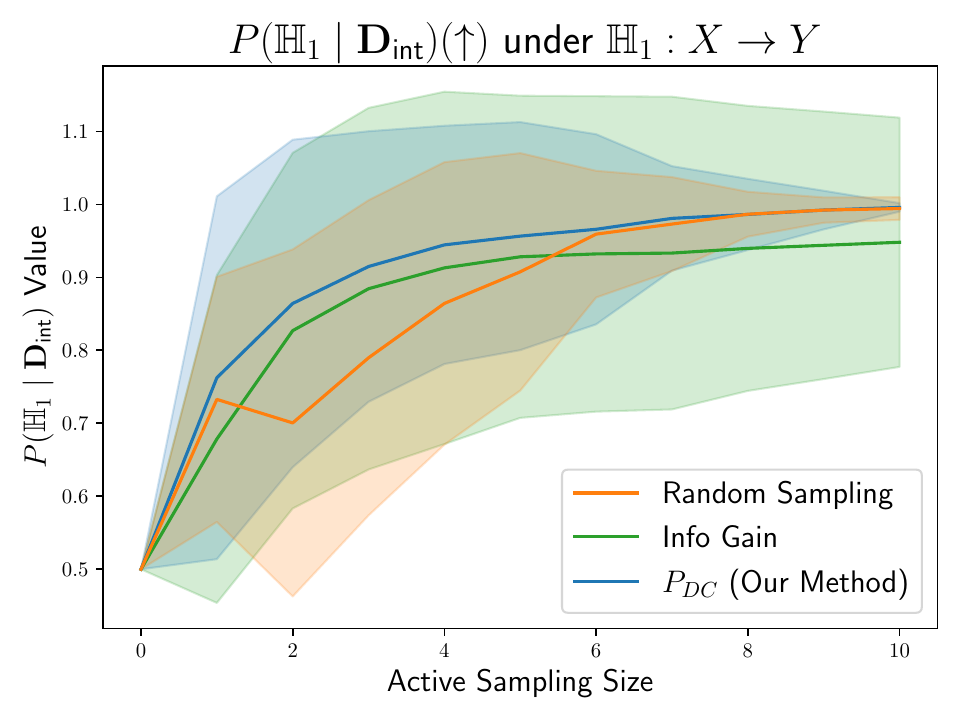}
    \label{fig:all_random_k030ph_gt_h1_x_to_y}
}
\caption{Results under different ground truths with $k_0 = \frac{1}{k_1} = 10$: $P_{DC}$, $\log \text{BF}_{01}$, and $P(\mathbb{H}_{gt} \mid \mathbf{D}_\text{int})$. But with random mean and random covariance compared to the settings in the Figure \ref{fig:k030results_h0_h1}. The first row corresponds to $\mathbb{H}_0$ ($X \gets Y$)). The second row corresponds to $\mathbb{H}_0$ ($X \gets U\to Y$), and the last row corresponds to $\mathbb{H}_1$ ($X \to Y$).}
\label{fig:all_random_k030results_h0_h1}
\end{center}
\vskip -0.2in
\end{figure}

\begin{figure}[ht]
\vskip 0.2in
\begin{center}
\subfigure{
    \includegraphics[width=0.3\columnwidth]{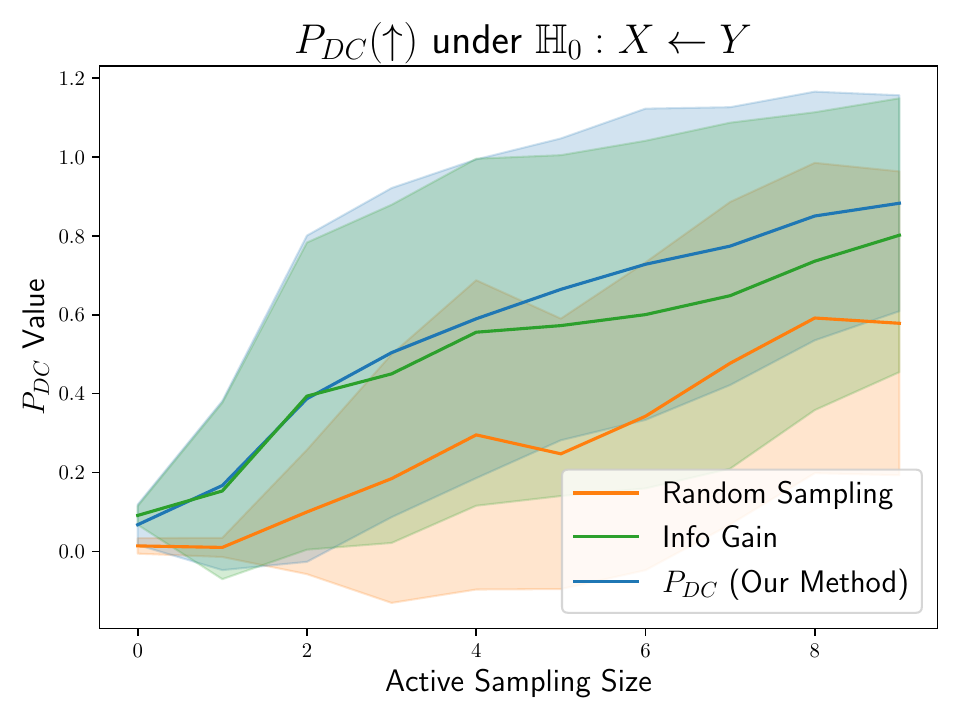}
    \label{fig:k0100pdc_h0_y_to_x}
}
\hfill
\subfigure{
    \includegraphics[width=0.3\columnwidth]{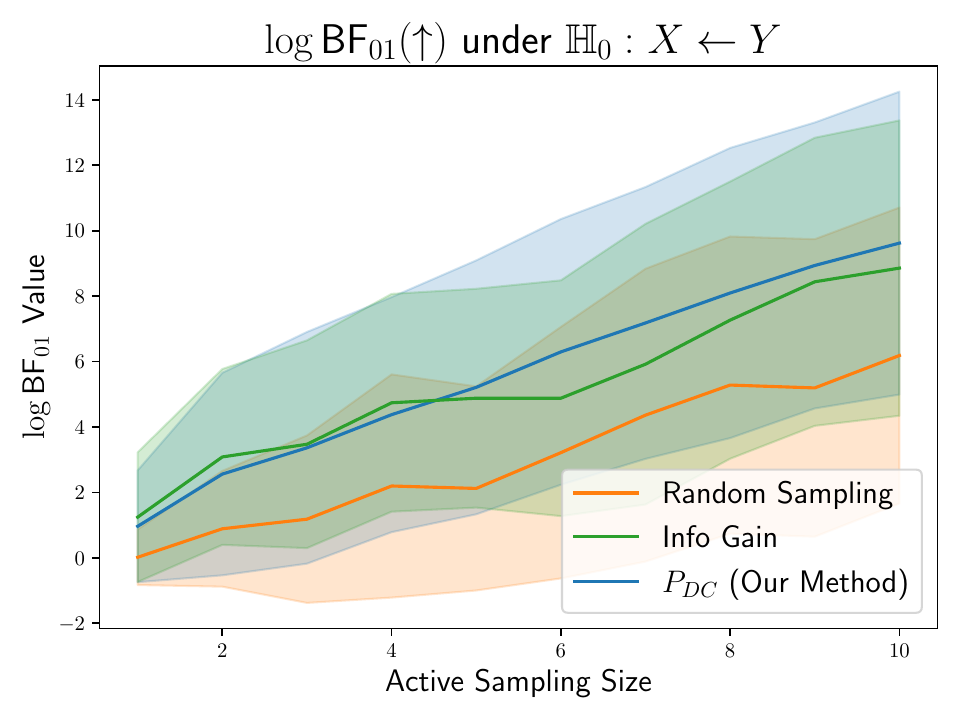}
    \label{fig:k0100log_bf_h0_y_to_x}
}
\hfill
\subfigure{
    \includegraphics[width=0.3\columnwidth]{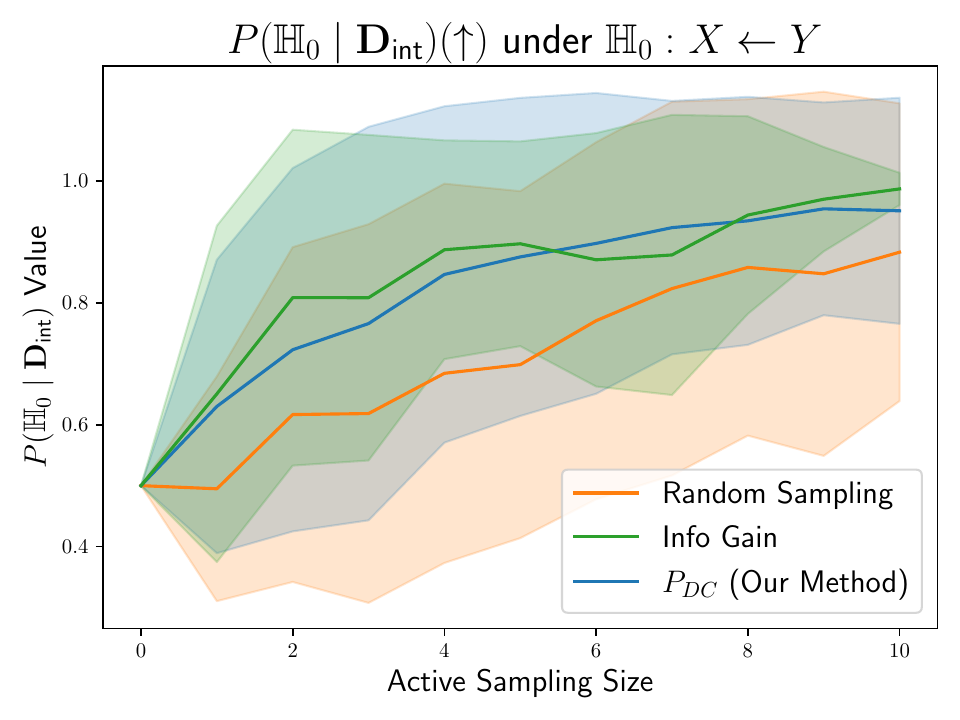}
    \label{fig:k0100ph_gt_h0_y_to_x}
}
\hfill
\subfigure{
    \includegraphics[width=0.3\columnwidth]{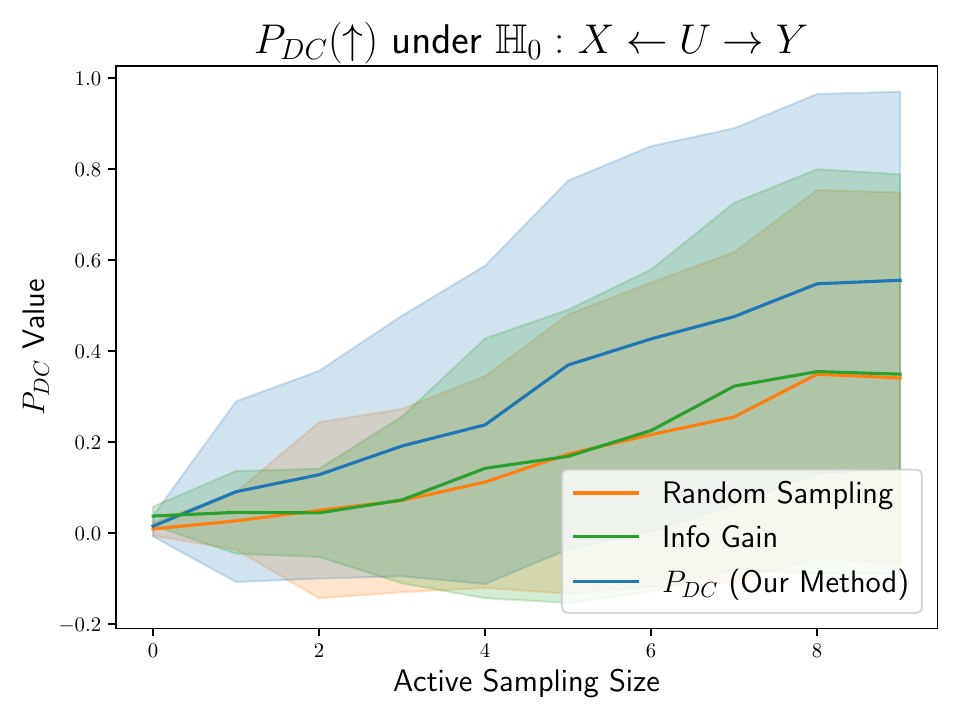}
    \label{fig:k0100pdc_h0_confounder}
}
 \hfill
\subfigure{
    \includegraphics[width=0.3\columnwidth]{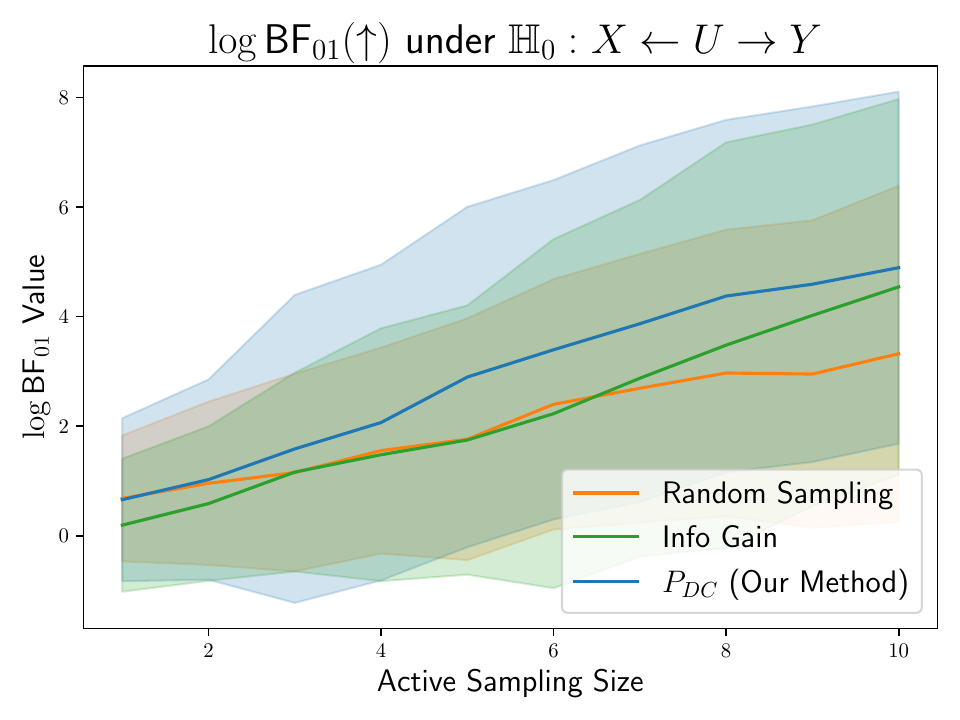}
    \label{fig:k0100log_bf_h0_confounder}
}
\hfill
\subfigure{
    \includegraphics[width=0.3\columnwidth]{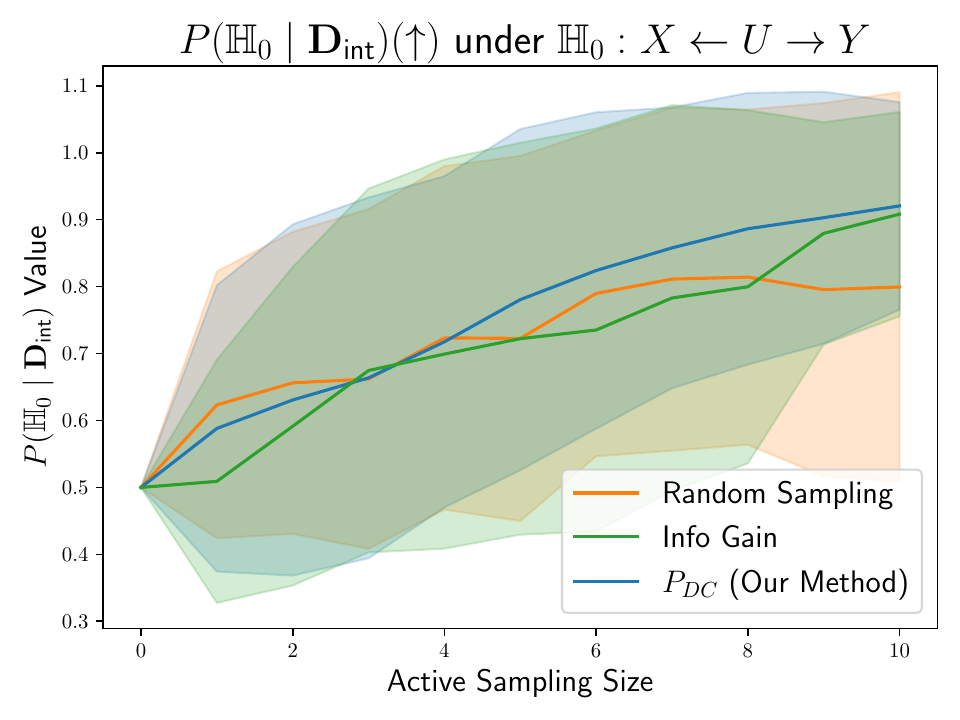}
    \label{fig:k0100ph_gt_h0_confounder}
}
\hfill
\subfigure{
    \includegraphics[width=0.3\columnwidth]{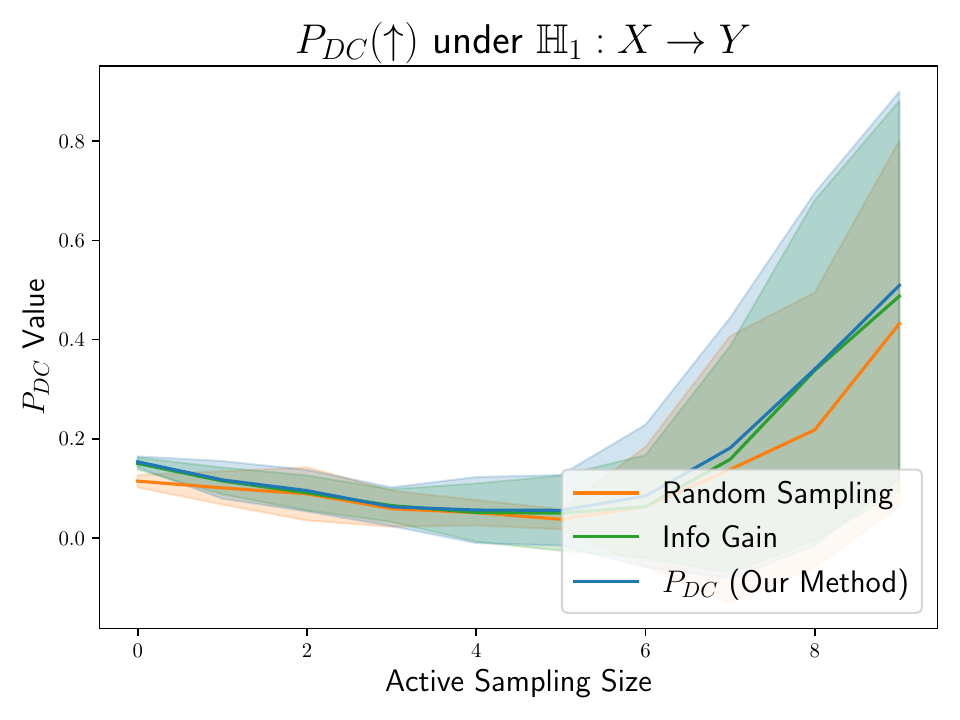}
    \label{fig:k0100pdc_h1_x_to_y}
}
\hfill
\subfigure{
    \includegraphics[width=0.3\columnwidth]{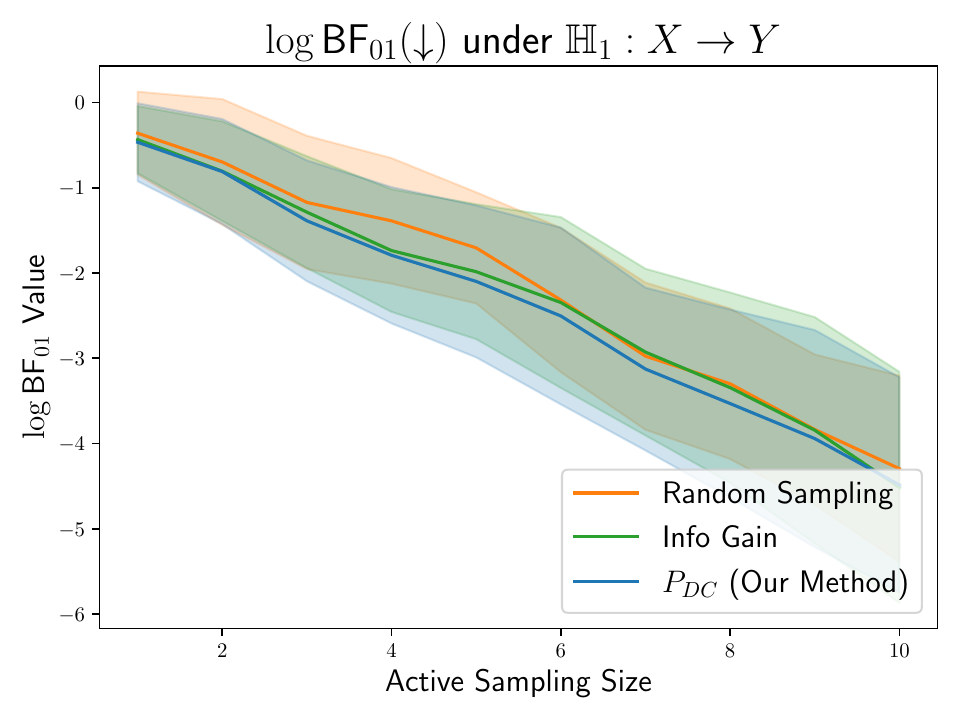}
    \label{fig:k0100log_bf_h1_x_to_y}
}
\hfill
\subfigure{
    \includegraphics[width=0.3\columnwidth]{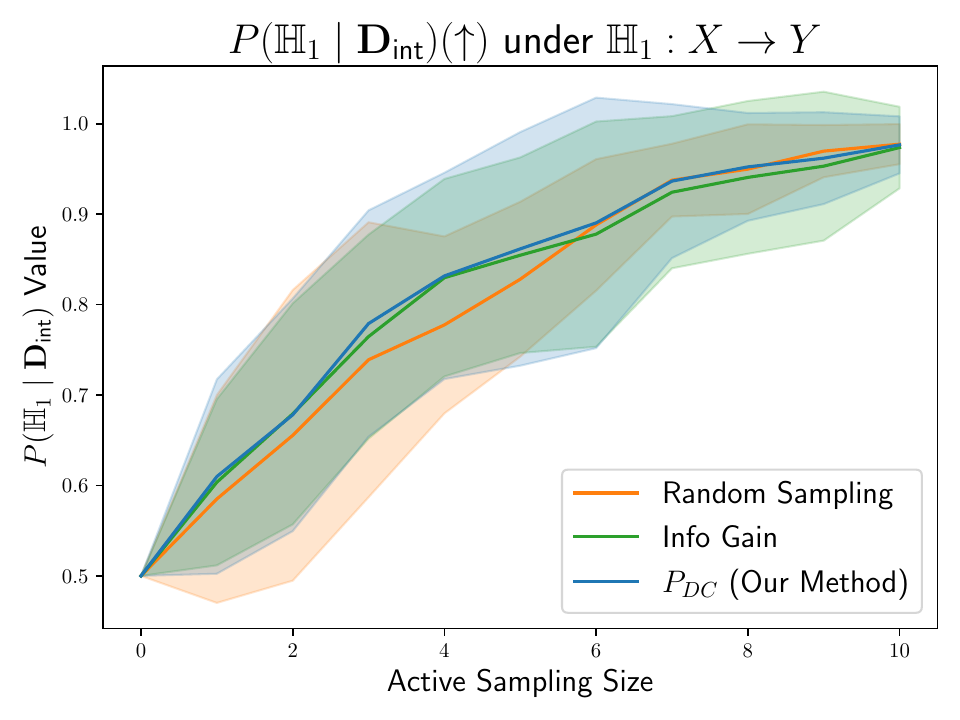}
    \label{fig:k0100ph_gt_h1_x_to_y}
}
\caption{Results under different ground truths with $k_0 = \frac{1}{k_1} = 10$: $P_{DC}$, $\log \text{BF}_{01}$, and $P(\mathbb{H}_{gt} \mid \mathbf{D}_\text{int})$. The first row corresponds to $\mathbb{H}_0$ ($X \gets Y$)). The second row corresponds to $\mathbb{H}_0$ ($X \gets U\to Y$), and the last row corresponds to $\mathbb{H}_1$ ($X \to Y$).}
\label{fig:k0100results_h0_h1}
\end{center}
\vskip -0.2in
\end{figure}

\begin{figure}[ht]
\vskip 0.2in
\begin{center}
\subfigure{
    \includegraphics[width=0.3\columnwidth]{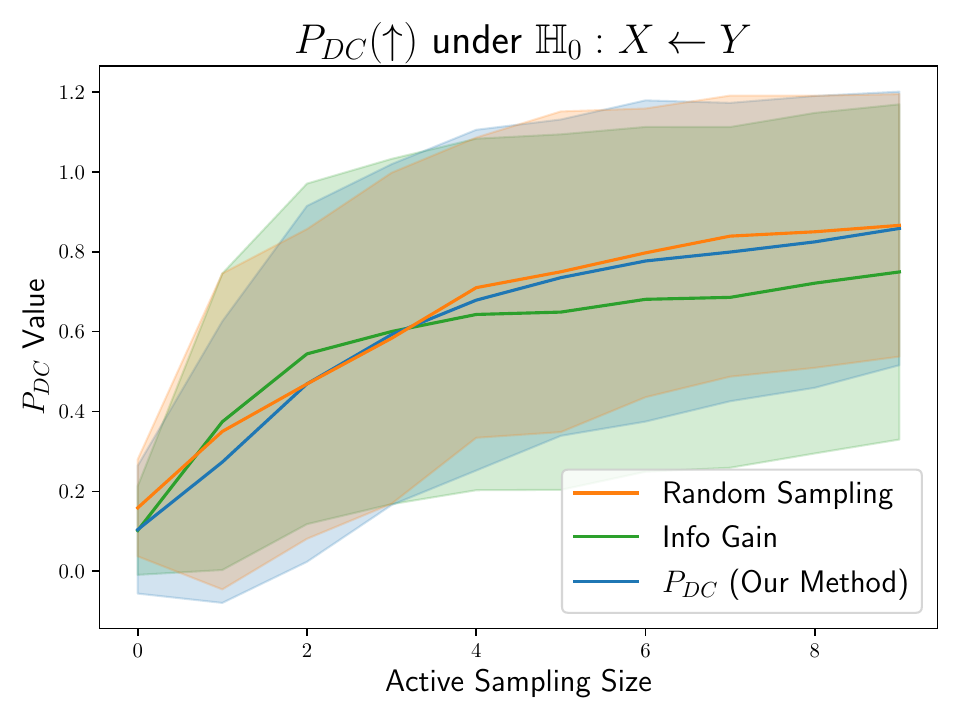}
    \label{fig:all_random_k0100pdc_h0_y_to_x}
}
\hfill
\subfigure{
    \includegraphics[width=0.3\columnwidth]{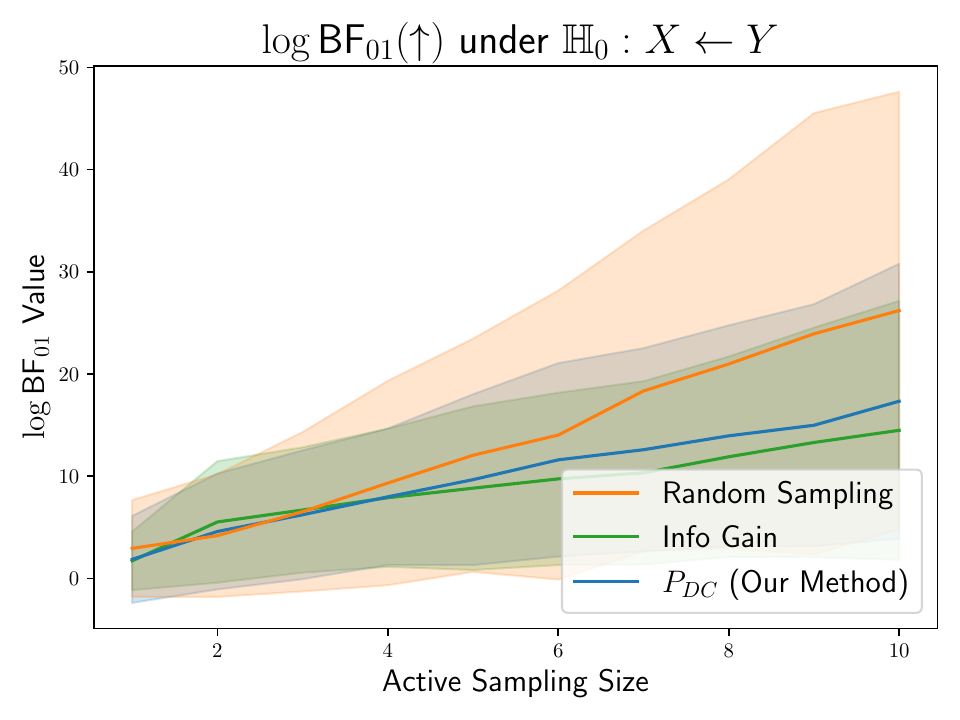}
    \label{fig:all_random_k0100log_bf_h0_y_to_x}
}
\hfill
\subfigure{
    \includegraphics[width=0.3\columnwidth]{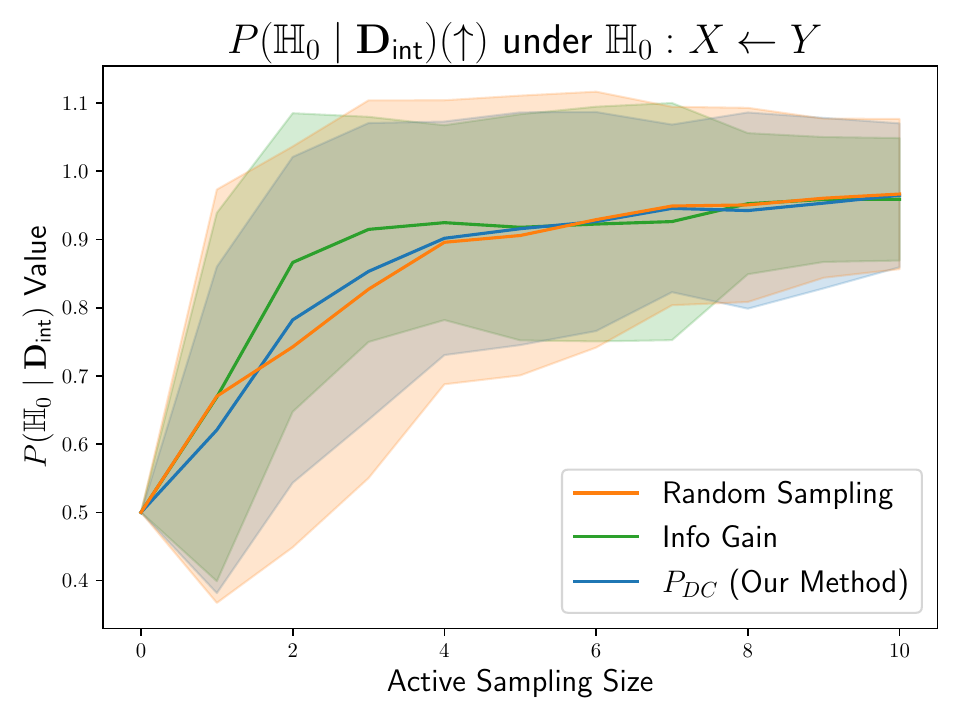}
    \label{fig:all_random_k0100ph_gt_h0_y_to_x}
}
\hfill
\subfigure{
    \includegraphics[width=0.3\columnwidth]{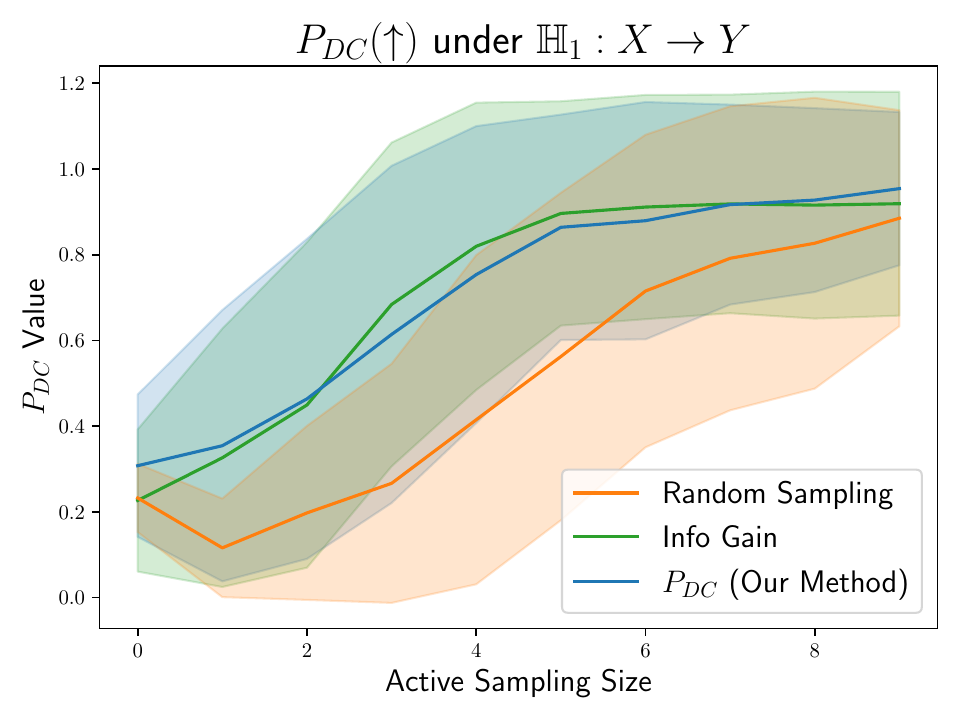}
    \label{fig:all_random_k0100pdc_h1_x_to_y}
}
\hfill
\subfigure{
    \includegraphics[width=0.3\columnwidth]{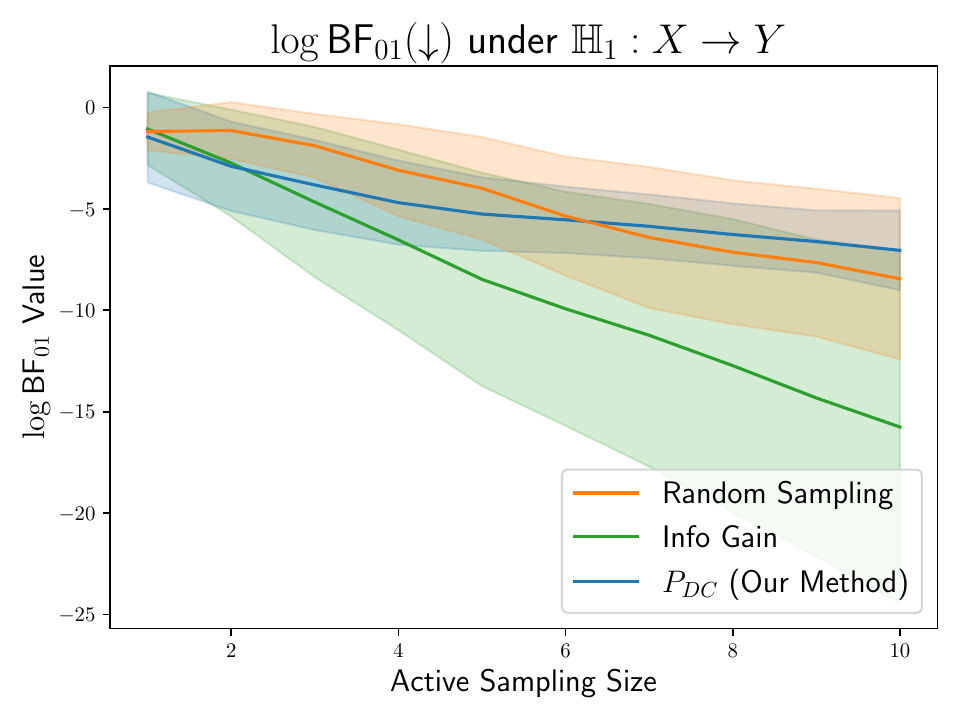}
    \label{fig:all_random_k0100log_bf_h1_x_to_y}
}
\hfill
\subfigure{
    \includegraphics[width=0.3\columnwidth]{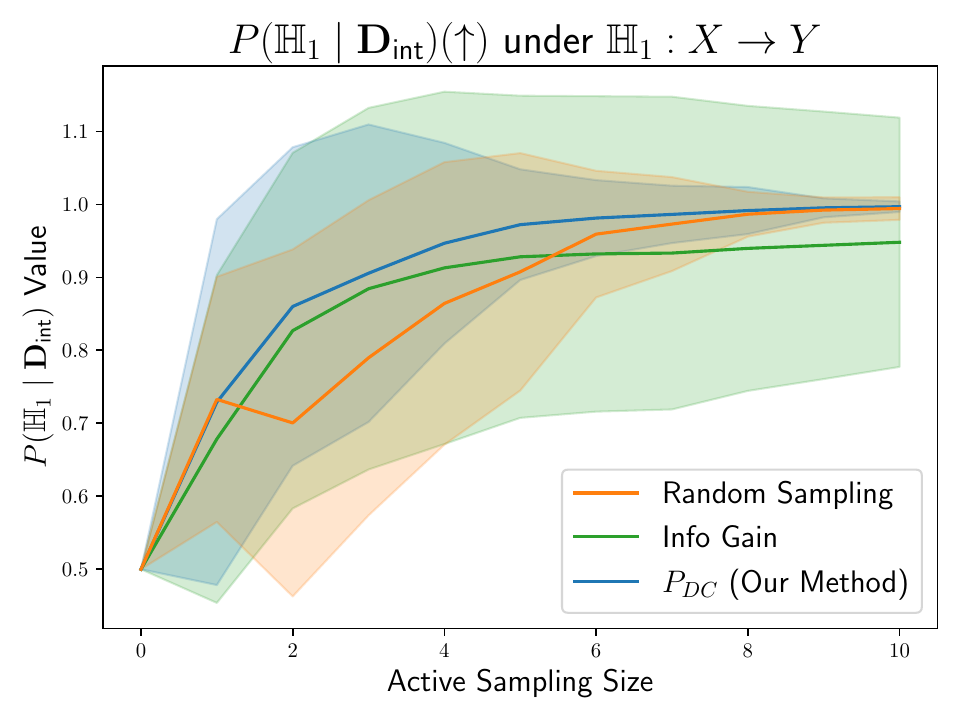}
    \label{fig:all_random_k0100ph_gt_h1_x_to_y}
}
\caption{Results under different ground truths with $k_0 = \frac{1}{k_1} = 10$: $P_{DC}$, $\log \text{BF}_{01}$, and $P(\mathbb{H}_{gt} \mid \mathbf{D}_\text{int})$. But with random mean and random covariance compared to the settings in the Figure \ref{fig:k0100results_h0_h1}. The first row corresponds to $\mathbb{H}_0$ ($X \gets Y$)). The second row corresponds to $\mathbb{H}_0$ ($X \gets U\to Y$), and the last row corresponds to $\mathbb{H}_1$ ($X \to Y$).}
\label{fig:all_random_k0100results_h0_h1}
\end{center}
\vskip -0.2in
\end{figure}